\documentclass[lettersize,journal]{IEEEtran}
\usepackage{amsmath,amsfonts}
\usepackage{algorithmic}
\usepackage{algorithm}
\usepackage{array}
\usepackage[caption=false,font=normalsize,labelfont=sf,textfont=sf]{subfig}
\usepackage{textcomp}
\usepackage{stfloats}
\usepackage{url}
\usepackage{verbatim}
\usepackage{graphicx}
\usepackage{cite}
\usepackage{amssymb}
\usepackage{xcolor}
\usepackage{enumitem}
\usepackage{adjustbox}
\usepackage{multirow}
\usepackage{balance}

\newcommand{\filledcirc}{\raisebox{0.5pt}{\scalebox{1.2}{$\bullet$}}}

\begin{document}

\title{How Does A Text Preprocessing Pipeline Affect Ontology Matching?}

\author{
\IEEEauthorblockN{
    Zhangcheng Qiang\IEEEauthorrefmark{1}\thanks{Corresponding author: Zhangcheng Qiang (qzc438@gmail.com)},
    Kerry Taylor\IEEEauthorrefmark{1},
    Weiqing Wang\IEEEauthorrefmark{2}
}\\
\IEEEauthorblockA{\IEEEauthorrefmark{1}Australian National University, Canberra, Australia}
\IEEEauthorblockA{\IEEEauthorrefmark{2}Monash University, Melbourne, Australia}\\
\texttt{qzc438@gmail.com}, \texttt{kerry.taylor@anu.edu.au, \texttt{teresa.wang@monash.edu}}
}

\maketitle

\begin{abstract}
The classical text preprocessing pipeline, comprising Tokenisation, Normalisation, Stop Words Removal, and Stemming/Lemmatisation, has been implemented in many systems for ontology matching (OM). However, the lack of standardisation in text preprocessing creates diversity in the mapping results. In this paper, we investigate the effect of the text preprocessing pipeline on 8 Ontology Alignment Evaluation Initiative (OAEI) tracks with 49 distinct alignments. We find that Tokenisation and Normalisation (categorised as Phase 1 text preprocessing) are more effective than Stop Words Removal and Stemming/Lemmatisation (categorised as Phase 2 text preprocessing). We propose two novel approaches to repair unwanted false mappings that occur in Phase 2 text preprocessing. One is a pre hoc logic-based repair approach used before text preprocessing, employing an ontology-specific check to find common words that cause false mappings. The other repair approach is the post hoc large language model (LLM)-based approach, used after text preprocessing, which utilises the strong background knowledge provided by LLMs to repair non-existent and counter-intuitive false mappings. The experimental results indicate that these two approaches can significantly improve the matching correctness and the overall matching performance.
\end{abstract}


\section{Introduction}

Ontology matching (OM) is crucial for enabling interoperability between heterogeneous ontologies, thereby facilitating the integration of disparate knowledge graphs (KGs)~\cite{grimm2007knowledge}. An OM process usually takes two ontologies as input, discovers mappings between entities, and produces a set of correspondences~\cite{euzenat2007ontology}. The classical text preprocessing pipeline is commonly used in OM. The pipeline consists of a set of steps to segment, reconstruct, analyse, and process text, namely Tokenisation (T), Normalisation (N), Stop Words Removal (R), and Stemming/Lemmatisation (S/L)~\cite{manning2008introduction}. T is the process of breaking the text into the smallest units (i.e. tokens). N transforms these different tokens into a single canonical form. R is the process of removing filler words that usually carry little meaning and can be omitted in most cases. S/L is used to deal with the grammatical variation of words, applying rules to find the simplest common form of the word.

\begin{figure}[htbp]
\centering
\includegraphics[width=0.8\columnwidth]{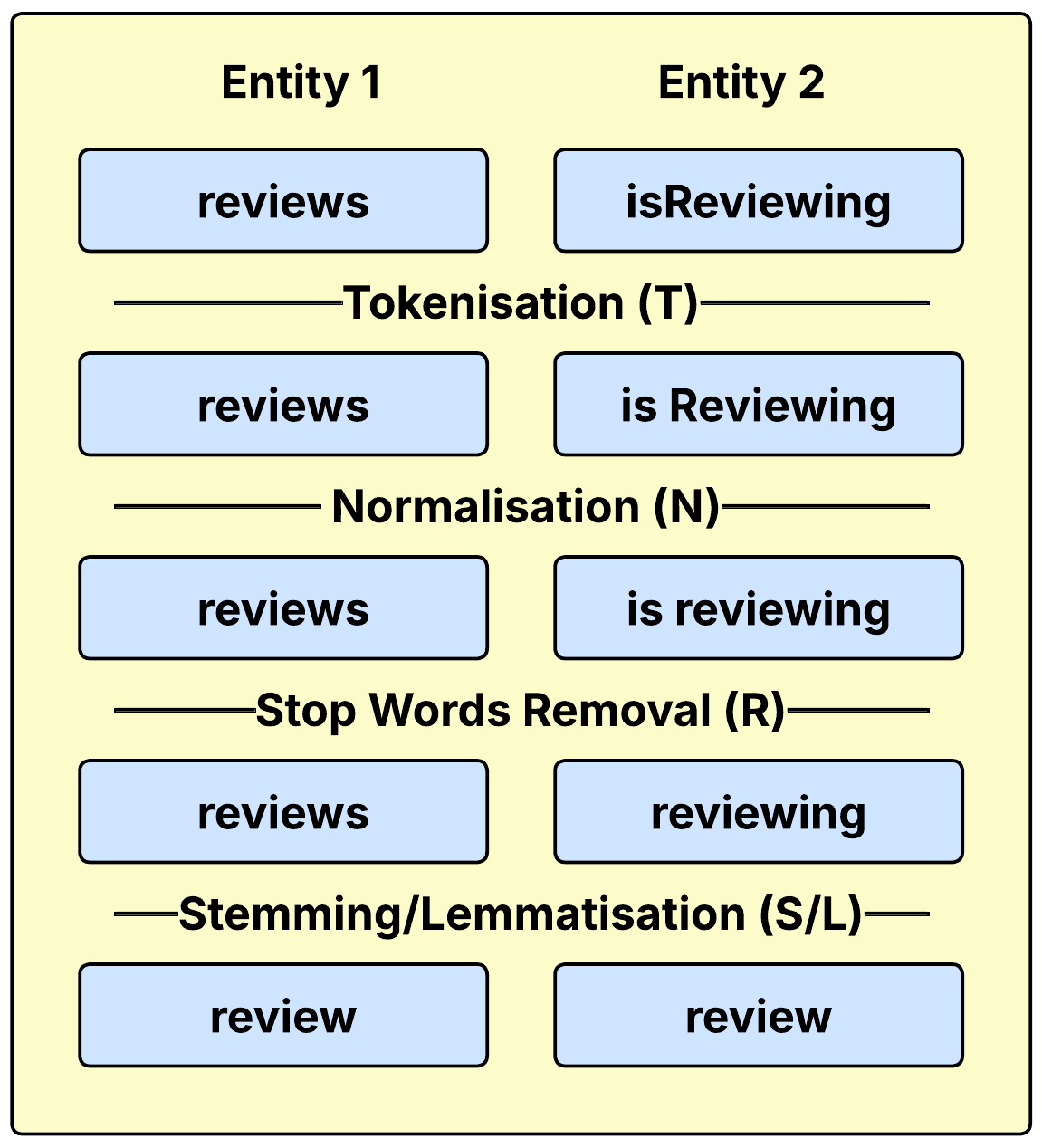}
\caption{An example of using the text preprocessing pipeline in OM.}
\label{fig: example}
\end{figure}

Fig.~\ref{fig: example} shows that two different ontology entities ``reviews'' and ``isReviewing'' can be matched via the text preprocessing pipeline as follows. After tokenisation, our tokens comprise a sequence of word units which we present as separated by white spaces within a token (``reviews'' \textit{cf} ``is Reviewing'). We then normalise all words to lowercase (``reviews'' \textit{cf} ``is reviewing''), remove the common stop words ``is'' (``reviews'' \textit{cf} ``reviewing''), and finally apply stemming/lemmatisation to find the simplest word form ``review'' from both ``reviews'' and ``reviewing'' (``review'' \textit{cf} ``review'').

A number of OM systems utilise the text preprocessing pipeline, but few studies have justified the choice of a specific text preprocessing method and systematically studied its role in optimising the matching performance. Our study fills this gap. We first conduct a comprehensive experimental analysis of the text preprocessing pipeline used in OM across a wide range of domains, aiming to explain the impact on OM of each step in the pipeline. In each step, text preprocessing methods are evaluated for correctness and completeness. We find that some steps of the text preprocessing pipeline are currently less effective. Based on this insight, we propose two novel pipeline repair approaches. These two methods utilise the logic of ontological representations and the power of large language models (LLMs), showing promising results for repairing weaknesses of the text preprocessing pipeline. Specifically, this paper makes the following contributions.
\begin{itemize}[wide, noitemsep, topsep=0pt, labelindent=0pt]
\item We comprehensively evaluate the performance of each step in the text preprocessing pipeline on 8 datasets of the Ontology Alignment Evaluation Initiative (OAEI)~\cite{oaei} with 49 distinct alignments. While there is a significant improvement using Tokenisation (T) and Normalisation (N), Stop Words Removal (R) and Stemming/Lemmatisation (S/L) are currently less effective. We find the issue is caused by inappropriate stop words removal, over-stemming, and over-lemmatisation. We also compare the performance of (1) Stemming and Lemmatisation, (2) different stemming methods including Porter, Snowball, and Lancaster, and (3) Lemmatisation with and without Part-of-Speech (POS) Tagging.
\item We propose two approaches to repair the less effective text preprocessing pipeline: (1) pre hoc logic-based repair and (2) post hoc LLM-based repair. To reduce the number of false mappings, the logic-based repair approach applies a pre hoc check based on ontology-specific settings before text preprocessing, while the LLM-based repair utilises LLMs as a post hoc validator after text preprocessing. We examine the performance of OM across different LLMs and prompt templates. We find the optimal pipeline to combine the traditional text processing pipeline with modern LLMs for OM. The experimental results indicate that these two repair methods can significantly improve the matching correctness and the overall matching performance.
\end{itemize}

The paper is organised as follows. Section~\ref{sec: background and related work} reviews the background of OM and the use of text preprocessing and LLMs in OM. Section~\ref{sec: analysis} analyses the text preprocessing pipeline used in OM. Section~\ref{sec: repair} proposes text preprocessing pipeline repair approaches and experimentally validates their performance. Section~\ref{sec: limitations} discusses the limitations of this work. Section~\ref{sec: conclusion} concludes the paper.

\section{Background and Related Work}
\label{sec: background and related work}

\subsection{Ontology Matching (OM)}

An ontology provides a formalised conceptualisation of domain knowledge~\cite{gruber1993translation}. The basic component of an ontology is the entity. Each entity is assigned a unique identifier (i.e. IRI) and represents a unique concept of the domain. Entities can be a concept (i.e. class), relation (i.e. property), or instance (i.e. individual)~\cite{noy2001ontology}. Ideally, the entity names are not duplicated within a single ontology to minimise confusion and overlap. It is recommended that ontologies representing similar domain knowledge use similar terminologies for entity names~\cite{fp000}.

OM aims to capture and align similar concepts that occur in different ontologies. Given a source ontology ($O_s$) and a target ontology ($O_t$), OM establishes mappings between pairs of entities drawn from each of the two ontologies. A correspondence (a.k.a. \emph{a mapping}) is defined as a 4-tuple $(e_1, e_2, r, c)$, where $e_1 \in O_s$ and $e_2 \in O_t$. $r$ is the relationship between two matched entities $e_1$ and $e_2$, and $c \in [0,1]$ is the confidence of the correspondence (or may be absent). The relationship $r$ in OM tasks can be equivalence ($\equiv$), subsumption ($\subseteq$), or other more complex relationships~\cite{euzenat2007ontology}. An alignment (A) is a set of candidate correspondences, while a reference (R) is a set of gold-standard correspondences verified by domain experts (i.e. the ground truth alignment).

A classical OM system usually contains syntactic matching, lexical matching, and semantic matching. This multilayer architecture has been implemented in several successful OM systems, including LogMap~\cite{jimenez2011logmap1,jimenez2011logmap2}, AgreementMakerLight (AML)~\cite{faria2013agreementmakerlight,faria2014agreementmakerlight}, FCA-Map~\cite{zhao2018matching,li2021combining}, and Agent-OM~\cite{qiang2023agent}. Syntactic matching is often the first step of classical OM. Text preprocessing is commonly used for syntactic matching. It has proven to correctly match many ontology entities because entities with similar string combinations are likely to be correct, and it is computationally efficient.

\subsection{Text preprocessing in OM}

The use of text preprocessing in OM can be traced back to the early stages of OM systems, which were initially developed and widely used as a basic component to generate linguistic-based mappings. AgreementMaker~\cite{cruz2009agreementmaker} has a normaliser to unify the textual information of the entities. SAMBO~\cite{lambrix2006sambo} uses the Porter Stemmer for each word to improve the similarity measure for terms with different prefixes and suffixes. In RiMOM~\cite{li2008rimom}, the context information of each entity is viewed as a document. The text in each document is preprocessed with tokenisation, stop words removal, and stemming. Recently, machine learning (ML) models have emerged for modern OM systems. While text preprocessing remains useful, its role is more focused on normalising the text that becomes the input to the model. For example, BERTMap~\cite{he2022bertmap} uses BERT's inherent WordPiece tokeniser to build the subword of each entity. DeepOnto~\cite{he2023deeponto} extends normalisation to axioms using EL embedding models~\cite{kulmanov2019embeddings}. The ML extension of LogMap~\cite{chen2021augmenting} reuses the seed mappings of traditional systems, where each entity is split into its component words and the mapping is based on pairwise component similarities.

To the best of our knowledge, most of the literature implements a preprocessing method without explaining why a specific method is chosen, and no studies have evaluated the effect of the entire text preprocessing pipeline.

\subsection{LLMs in OM}

Advances in LLM research raise the question of whether text preprocessing is obsolete for OM. Early explorations in~\cite{he2023exploring,norouzi2023conversational} show the potential of using LLMs for OM, but matching by LLM prompting alone commonly results in poor performance. OLaLa~\cite{hertling2023olala} proposes a prototype and states that the LLMs for OM include a number of decisions on model and parameter combinations, and there is no one-model-fits-all solution. Works on both Agent-OM~\cite{qiang2023agent} and LLM4OM~\cite{giglou2024llms4om} point out that LLMs have limited context length and inherited hallucinations. While LLM4OM highlights the importance of integrating the Retrieval-Augmented Generation (RAG)~\cite{lewis2021retrievalaugmented}, Agent-OM introduces an agent-based framework and enhances the compatibility of LLMs via function calling~\cite{openai-function-calling} for OM to improve matching performance. The authors in~\cite{taboada2025ontology} state that the use of LLMs for OM has a high computational overhead and apply a prioritised depth-first search (PDFS) strategy on the retrieve-then-prompt pipeline to optimise candidate selection. Recent works in~\cite{amini2024towards, zamazal2024towards,sousa2025complex} also utilise LLMs and LLM embeddings for complex OM.

However, these works focus on building a sophisticated framework for using LLMs for OM, and less attention has been paid to the subtask of using LLMs for text preprocessing as used in OM. The challenges are twofold. On one hand, LLM hallucinations may produce incorrect ``seed mappings''. Such mappings are toxic for the subsequent lexical and semantic matching. On the other hand, it is difficult to select a generic LLM prompting strategy from a number of LLM prompting templates and strategies. With strong background knowledge, LLMs can be helpful in detecting false mappings after text preprocessing. We discuss LLM-based repair in Section~\ref{sub-sec: llm-based-repair}.

\section{Text Preprocessing Pipeline Analysis}
\label{sec: analysis}

\subsection{Experiment Setup}

\begin{figure}[htbp]
\includegraphics[width=\columnwidth]{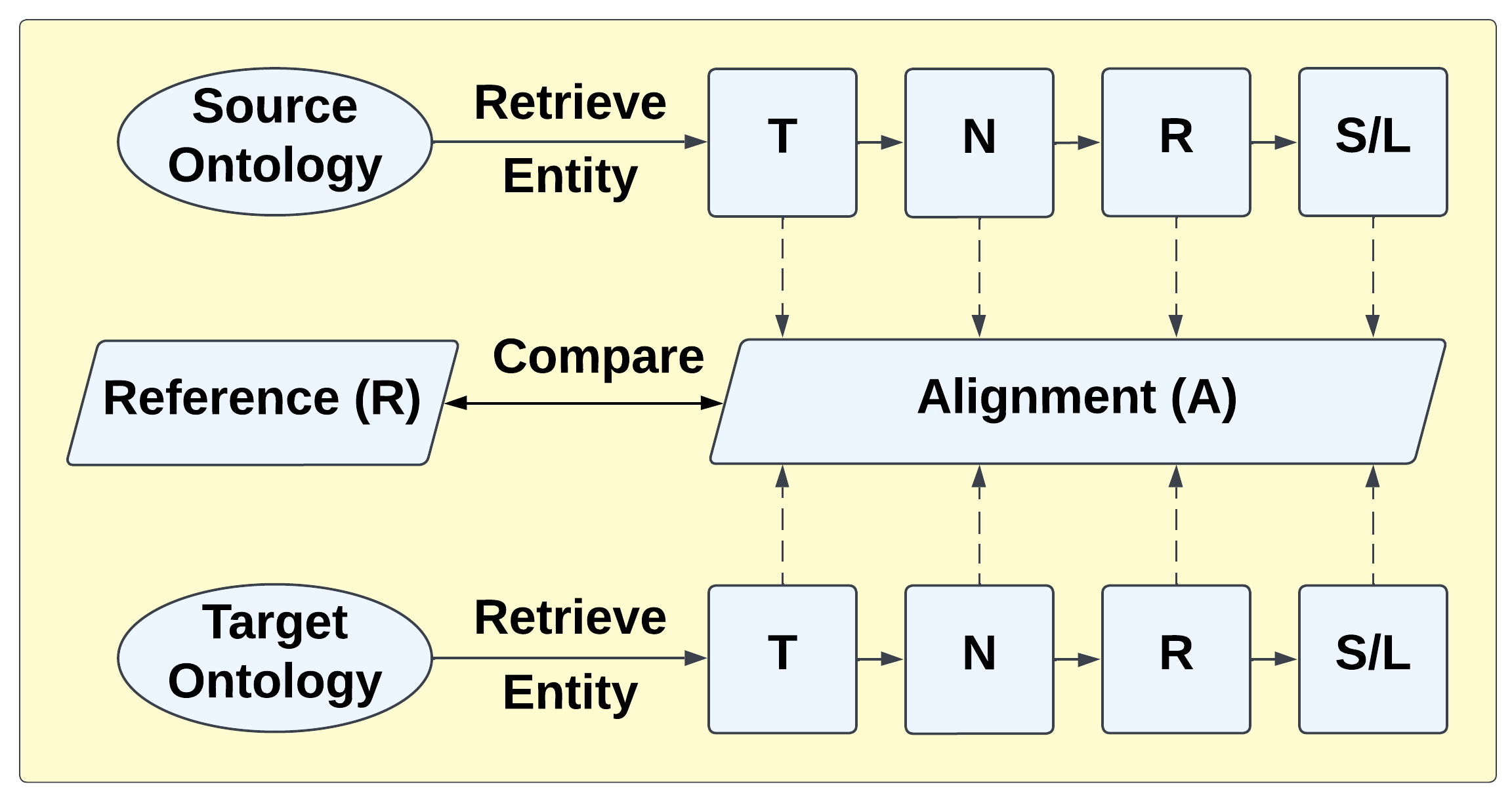}
\caption{Experiment setup to analyse the effect of text preprocessing pipeline in OM: Tokenisation (T), Normalisation (N), Stop Words Removal (R), and Stemming/Lemmatisation (S/L).}
\label{fig: methodology}
\end{figure}

Fig.~\ref{fig: methodology} shows the experiment setup to analyse the text preprocessing pipeline. Alignment (A) is generated via the text preprocessing pipeline. For both $O_s$ and $O_t$, we retrieve the entities from classes (owl:Class) and properties (both owl:ObjectProperty and owl:DatatypeProperty). For those ontologies where the names of the concepts are not textual (e.g. a code or numerical identifier), we retrieve the meaningful text from entity labels (rdfs:label or other annotation properties) instead. Then, we apply the text preprocessing pipeline method $f(\cdot)$ on each entity $e_1 \in O_s$ and $e_2 \in O_t$. If $f(e_1) = f(e_2)$, we store the correspondence in the corresponding alignment file. While OM systems often follow text preprocessing with a more sophisticated strategy to determine the equivalence of entities, we define equivalence to mean identical text generated by preprocessing alone. This enables us to directly evaluate the preprocessing performance as a standalone contribution to OM. We compare the generated alignment (A) with the reference (R) to evaluate the performance of the text preprocessing pipeline on OM.

\begin{figure}[htbp]
\centering
\includegraphics[width=1\columnwidth]{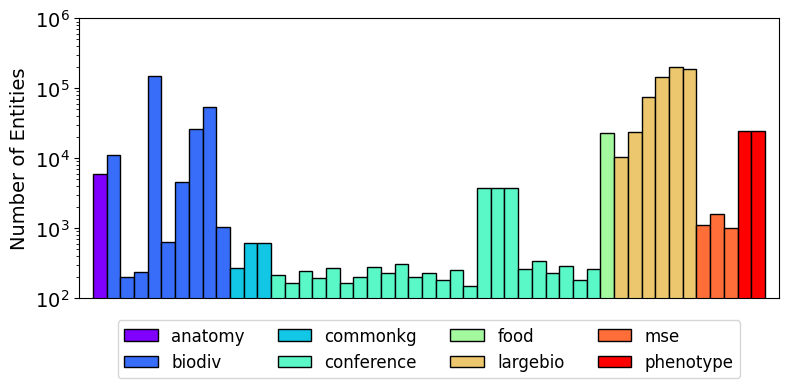}
\caption{Number of entities in each alignment across different tracks.}
\label{fig: metadata}
\end{figure}

We select 8 OAEI tracks stored in the Ontology Matching Evaluation Toolkit (MELT)~\cite{hertling2019melt} public repository (retrieved January 1, 2025). Table~\ref{tab: category of OAEI tracks} shows the details of the selected tracks. Each track may contain more than one alignment corresponding to different pairs of ontologies. For example, the {\itshape Largebio} Track has multiple reference files, pairing FMA~\cite{rosse2003reference} and NCI~\cite{golbeck2003national}, FMA and SNOMED~\cite{donnelly2006snomed}, and SNOMED and NCI, respectively. The number of alignments for each track is given in the table. There are 49 distinct alignments evaluated in this study. Fig.~\ref{fig: metadata} shows the number of entities in each alignment across different tracks.

\begin{table}[htbp]
\renewcommand\arraystretch{1.15}
\tabcolsep=0.15cm
\caption{Selected OAEI tracks and the number of alignments.}
\label{tab: category of OAEI tracks}
\begin{adjustbox}{width=1\columnwidth, center}
\begin{tabular}{|c|c|c|}
\hline
\multirow{1}{*}{\textbf{Name}}      & \multirow{1}{*}{\textbf{Domain}}  & \multirow{1}{*}{\textbf{\# Alignments}}             \\ \hline
\multicolumn{1}{|c|}{Anatomy}       & Human and Mouse Anatomy           & 1                                                   \\ \hline
\multicolumn{1}{|c|}{Biodiv}        & Biodiversity and Ecology          & 9                                                   \\ \hline
\multicolumn{1}{|c|}{CommonKG}      & Common Knowledge Graphs           & 3                                                   \\ \hline
\multicolumn{1}{|c|}{Conference}    & Conference                        & 24                                                  \\ \hline
\multicolumn{1}{|c|}{Food}          & Food Nutritional Composition      & 1                                                   \\ \hline
\multicolumn{1}{|c|}{Largebio}      & Biomedical                        & 6                                                   \\ \hline
\multicolumn{1}{|c|}{MSE}           & Materials Science \& Engineering  & 3                                                   \\ \hline
\multicolumn{1}{|c|}{Phenotype}     & Disease and Phenotype             & 2                                                   \\ \hline
\end{tabular}
\end{adjustbox}
\end{table}

\begin{figure}[htbp]
\centering
\subfloat[Classes]{\includegraphics[width=0.5\columnwidth]{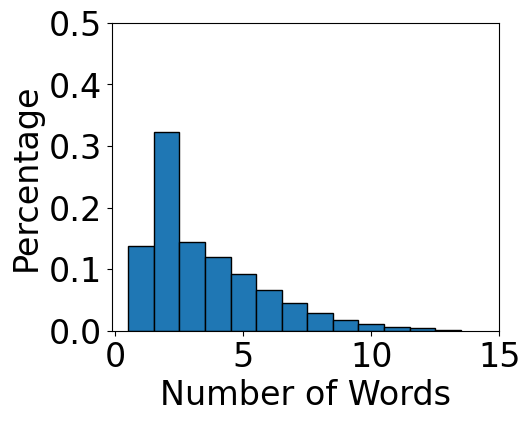}}
\subfloat[Properties]{\includegraphics[width=0.5\columnwidth]{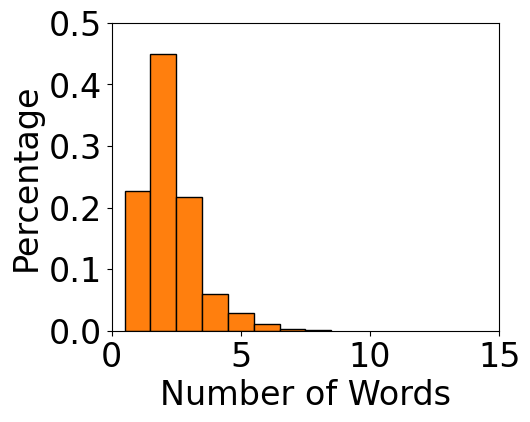}}
\caption{Frequency distribution of compound words. We exclude entities with more than 15 compound words because their proportion is less than 1\%.}
\label{fig: compound word}
\end{figure}

Compound words are frequently used in ontology naming conventions. For example, compound words such as ``art gallery'' can be formatted as the Pascal case ``ArtGallery'' used in YAGO~\cite{suchanek2007yago} or the Snake case ``art\_gallery'' used in Wikidata~\cite{vrandevcic2014wikidata}. Fig.~\ref{fig: compound word} shows the frequency distribution of compound words in the selected OAEI tracks. For comparisons between entities with compound words, one approach is to use their tokens, where ``XY'' is equivalent to ``YX'' because they share the same concepts X and Y. However, our comparison considers both concepts and their order, so that ``XY'' is not equivalent to ``YX''. Although concurrent ``XY'' and ``YX'' are rare in the matching process, we assume that it could happen and so we take account of the order of concepts in compound word names to ensure a fair comparison in our experiments.

There are a variety of methods that can be employed in a general text preprocessing pipeline, but not all of them are applicable to syntactic OM. We select the following methods in each text preprocessing step: (a) Tokenisation (T): We use the conventional full word tokenisation method to separate the compound words in an entity name. Experiments on several alternative \emph{subword} tokenisation methods (e.g. Byte Pair Encoding~\cite{gage1994new,sennrich2015neural}, WordPiece~\cite{schuster2012japanese,devlin2019bert}, and Unigram~\cite{kudo2018subword,kudo2018sentencepiece}) show that results after concatenating the processed name parts back together are generally the same as for full word tokenisation. We do not use sentence tokenisation (a.k.a. segmentation) because text retrieved from the entity's name or label is commonly short. (b) Normalisation (N): includes lowercasing, HTML tag removal, separator formatting, and punctuation removal. Other methods that may potentially change the word semantics are excluded (e.g. removal of special characters and numbers). (c) Stop Words Removal (R): includes the most common English stop words defined in the Natural Language Toolkit (NLTK)~\cite{bird2008multidisciplinary}. (d) Stemming/Lemmatisation (S/L): Stemming methods include Porter, Snowball, and Lancaster. Lemmatisation utilises the NLTK Lemmatiser based on WordNet~\cite{miller1995wordnet}, and word categorisation employs POS Tagging.

\begin{figure}[htbp]
\centering
\begin{minipage}{0.49\columnwidth}
\centering
\includegraphics[width=4cm,height=3cm]{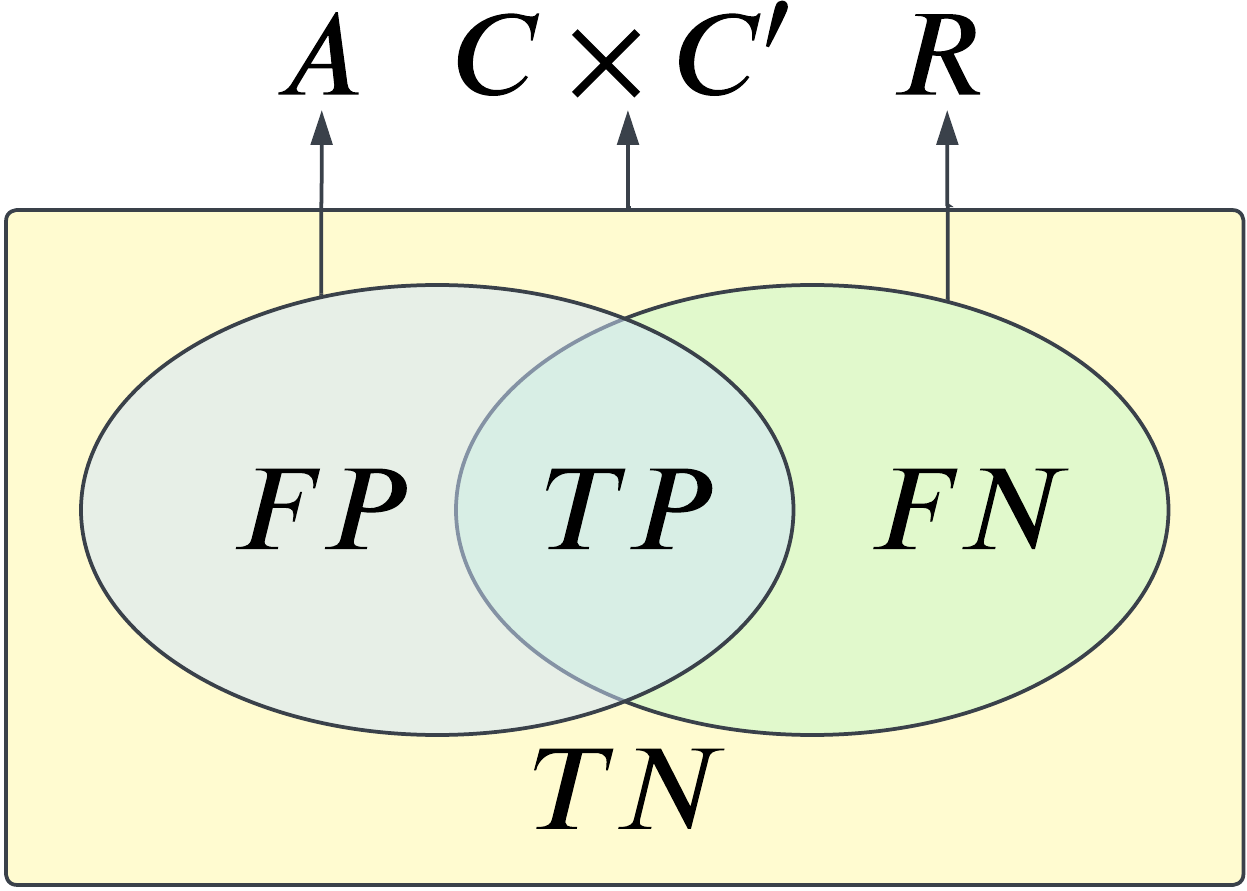}
\caption{OM evaluation~\cite{euzenat2007semantic}.}
\label{fig: om-evaluation}
\end{minipage}
\hfill
\begin{minipage}{0.49\columnwidth}
\begin{equation*}
\label{eq:ontology_matching}
\begin{aligned}
Precision (Prec) = \frac{|A \cap R|}{|A|}
\\
Recall (Rec) = \frac{|A \cap R|}{|R|}
\\
F_1 \ Score = \frac{2}{Prec^{-1} + Rec^{-1}}
\end{aligned}
\end{equation*}
\end{minipage}
\end{figure}

There are four primitive measures in information retrieval: true positive (TP), false positive (FP), false negative (FN), and true negative (TN). In the context of OM, the evaluation compares an alignment (A) returned by the OM system with a gold standard reference (R) validated by domain experts. Fig.~\ref{fig: om-evaluation} illustrates that the four primitive measures in OM can be interpreted as $TP = A \cap R$, $FP = A-R$, $FN = R-A$, and $TN = (C \times C') - (A \cup R)$, where $C \times C'$ refers to all possible correspondences. Accuracy (Acc), Specificity (Spec), Precision (Prec), Recall (Rec), and $F_\beta$ Score ($F_\beta$) are the most common evaluation measures based on TP, FP, FN, and TN. In the context of OM, since $C \times C'$ is extremely large (the Cartesian product of $e_1 \in O_s$ and $e_2 \in O_t$), TN is often much larger than TP, FP, and FN. This means that Accuracy (Acc) and Specificity (Spec) are close to 1, and they have no statistically significant difference across different alignments. We note that Precision (Prec) and Recall (Rec) contribute equally to $F_\beta$. Therefore, we choose Precision (Prec), Recall (Rec), and F1 Score ($\beta=1$) in this study. They are defined as shown in Fig.~\ref{fig: om-evaluation}.

\subsection{Results}
\label{sec: results}

\begin{figure}[!t]
\centering
\subfloat{
\includegraphics[width=0.325\columnwidth]{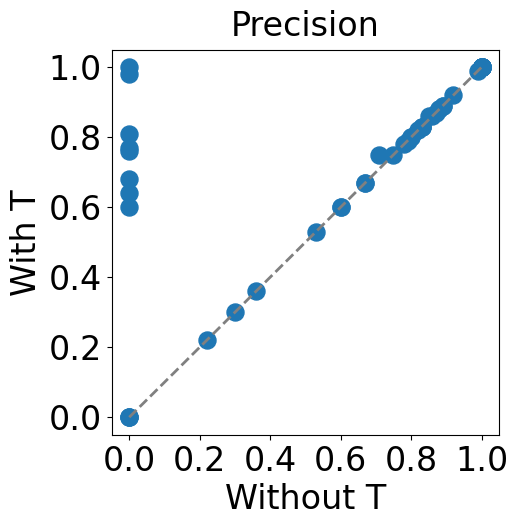}
\includegraphics[width=0.325\columnwidth]{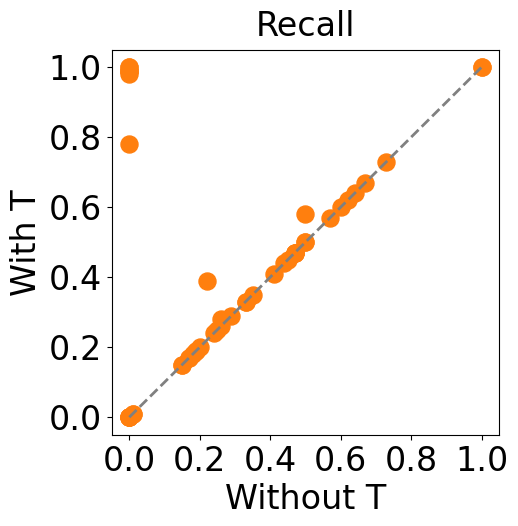}
\includegraphics[width=0.325\columnwidth]{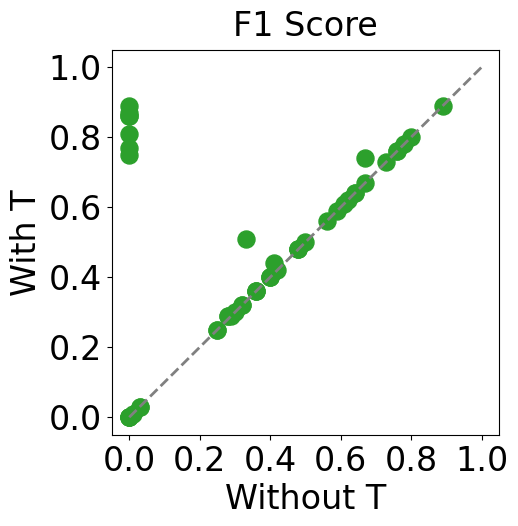}
}
\hfill
\subfloat{
\includegraphics[width=0.325\columnwidth]{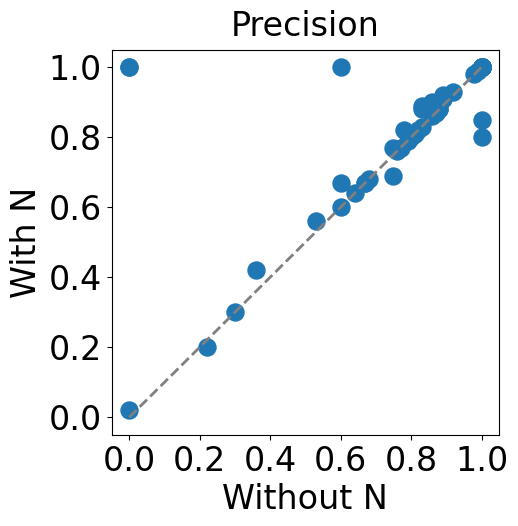}
\includegraphics[width=0.325\columnwidth]{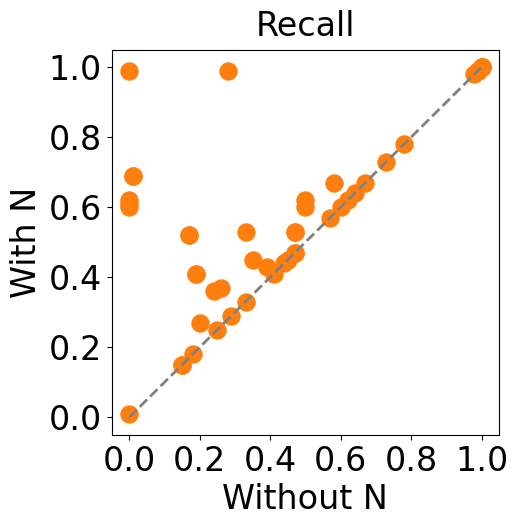}
\includegraphics[width=0.325\columnwidth]{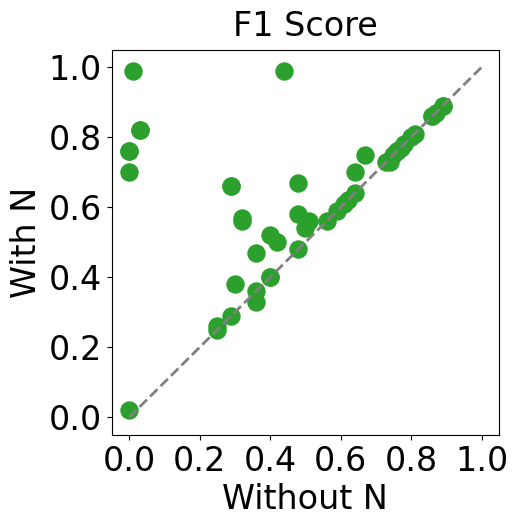}
}
\hfill
\subfloat{
\includegraphics[width=0.325\columnwidth]{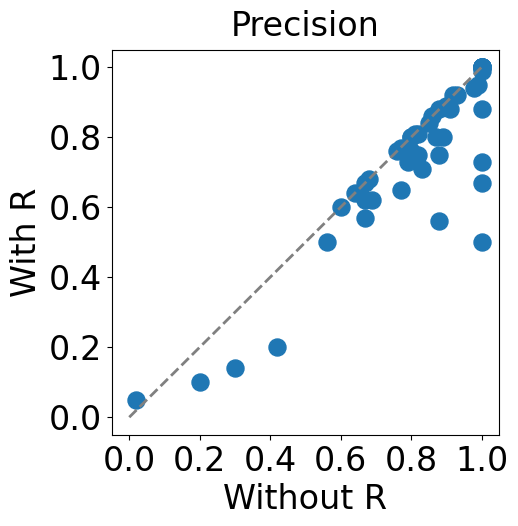}
\includegraphics[width=0.325\columnwidth]{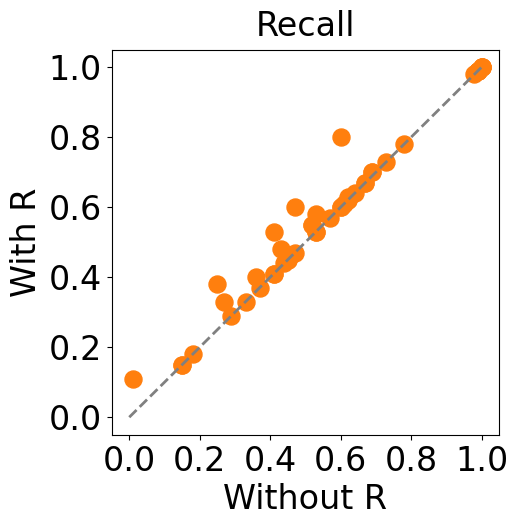}
\includegraphics[width=0.325\columnwidth]{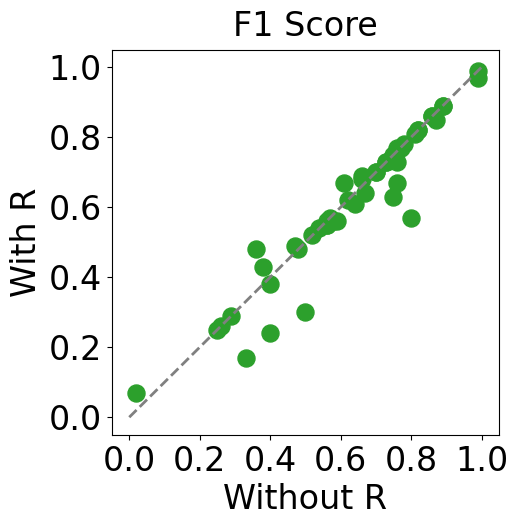}
}
\hfill
\subfloat{
\includegraphics[width=0.325\columnwidth]{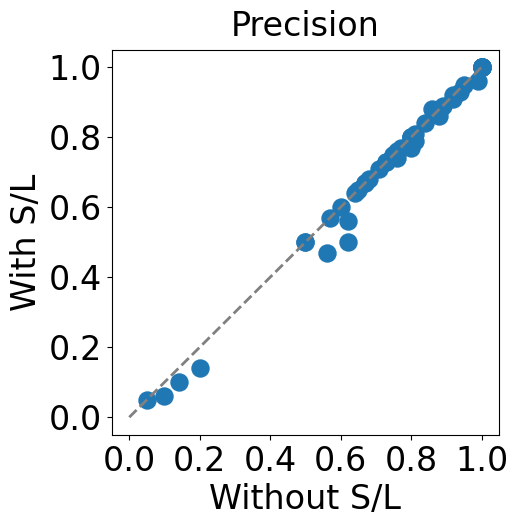}
\includegraphics[width=0.325\columnwidth]{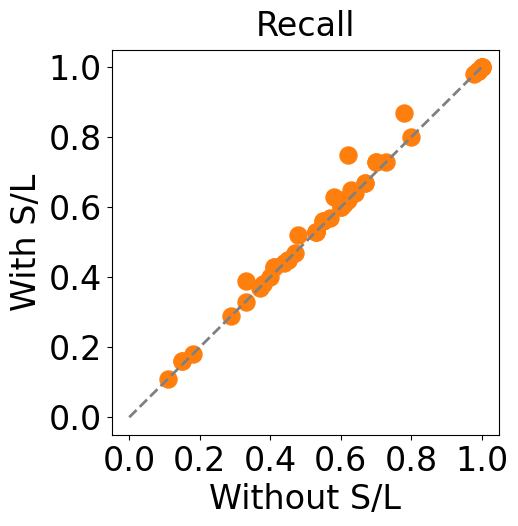}
\includegraphics[width=0.325\columnwidth]{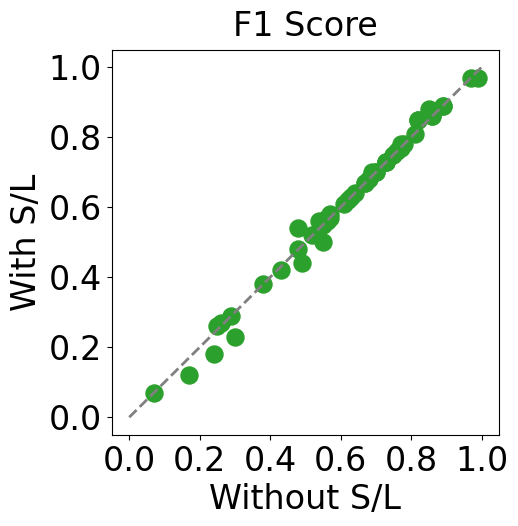}
}
\hfill
\subfloat{
\includegraphics[width=0.325\columnwidth]{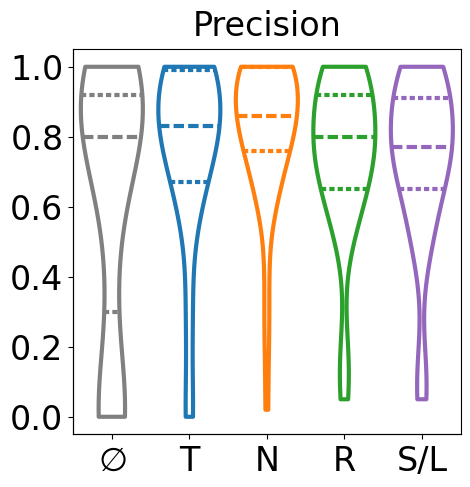}
\includegraphics[width=0.325\columnwidth]{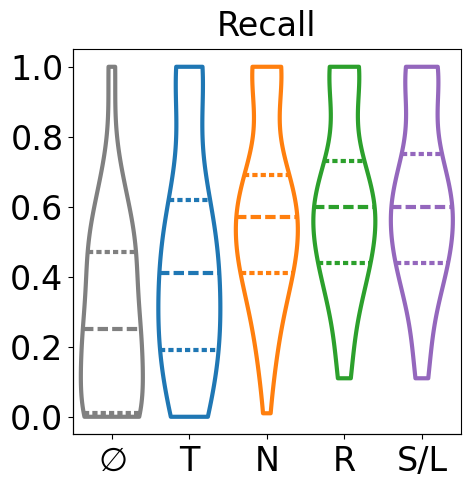}
\includegraphics[width=0.325\columnwidth]{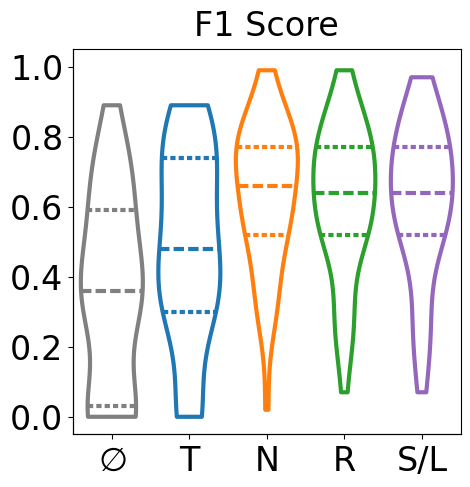}
}
\caption{Comparison of text preprocessing pipeline: Tokenisation (T), Normalisation (N), Stop Words Removal (R), and Stemming/Lemmatisation (S/L). In the violin plots, three horizontal lines inside each violin plot show three quartiles: Q1, median, and Q3 (an artefact of the Seaborn~\cite{waskom2021seaborn} package).}
\label{fig: compare-pipeline}
\end{figure}

Fig.~\ref{fig: compare-pipeline} compares each step of the text preprocessing pipeline. These steps are always applied sequentially in the pipeline. Beginning with $\varnothing$ (no text preprocessing), we see the performance of subsequent Tokenisation (T), Normalisation (N), Stop Words Removal (R), and Stemming/Lemmatisation (S/L). For T and N, most of the data points above the equivalence line in precision, recall, and F1 score indicate that they can help OM. For R and S/L, some data points are above the equivalence line in recall, but most data points are below the equivalence line in precision and F1 score indicating that they do not play a positive role. In the violin plots: (a) Precision: The median increases with T and N but decreases with R and S/L. After T, the shape of the distribution is unchanged by R and S/L. (b) Recall: After T, the median increases slightly with each of N, R, and S/L. The shape of the distribution does not change after N. (c) F1 Score: The median increases with T and N but then decreases with R and S/L. The shape of the distribution remains unchanged after N, which also indicates that R and S/L do not help OM.

\begin{figure}[!t]
\centering
\subfloat{
\includegraphics[width=0.325\columnwidth]{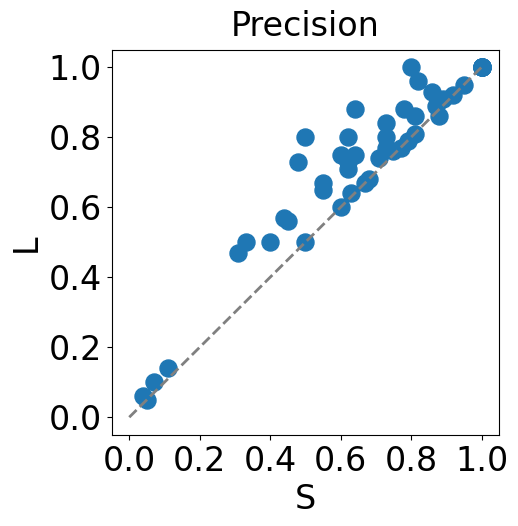}
\includegraphics[width=0.325\columnwidth]{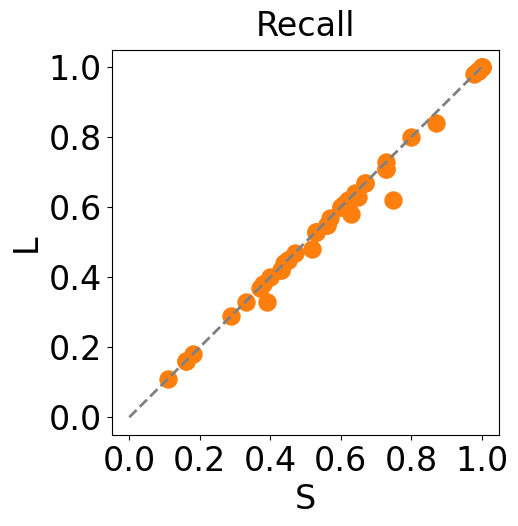}
\includegraphics[width=0.325\columnwidth]{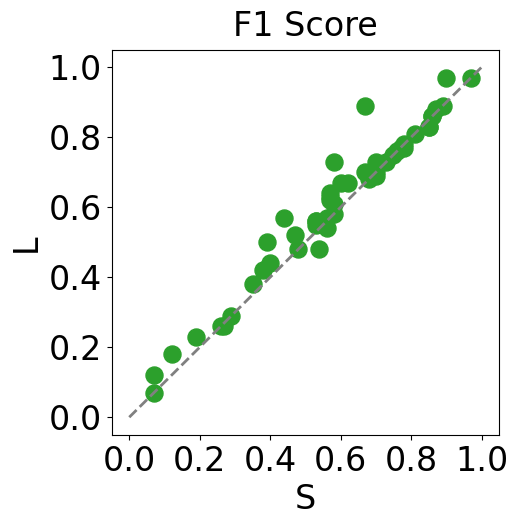}
}
\hfill
\subfloat{
\includegraphics[width=0.325\columnwidth]{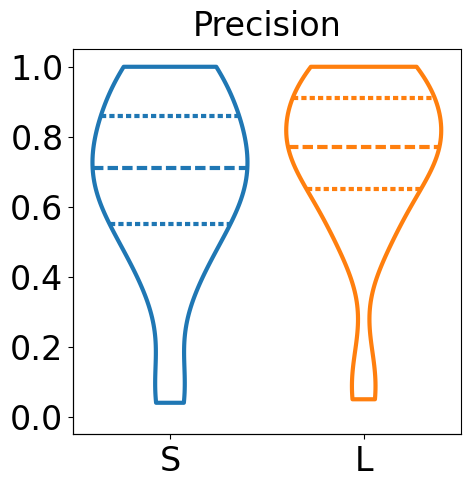}
\includegraphics[width=0.325\columnwidth]{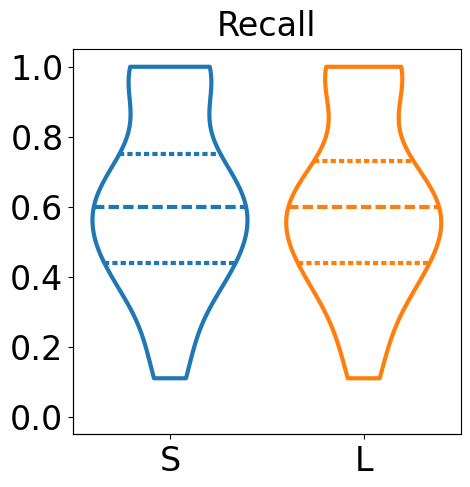}
\includegraphics[width=0.325\columnwidth]{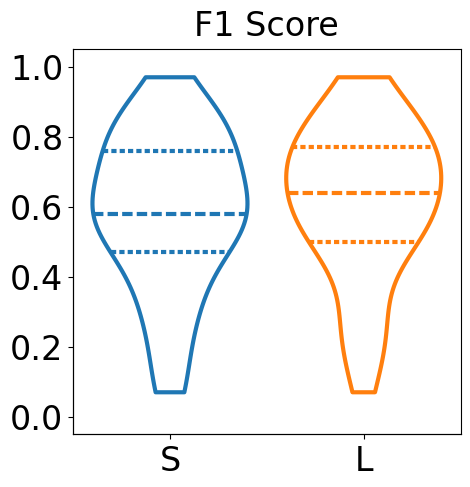}
}
\caption{Comparison of Stemming (S) and Lemmatisation (L).}
\label{fig: compare-sl}
\end{figure}

Fig.~\ref{fig: compare-sl} compares Stemming (S) and Lemmatisation (L) after Tokenisation (T) and Normalisation (N) have been applied. Most data points above the L = S line in precision and F1 score indicate that L is better than S in OM (assuming that post hoc correction is excluded). In the violin plots: (a) Precision: The median after L is greater than that achieved by S. The shape of the distribution is slightly different. (b) Recall: The median and the shape of the distribution are identical after S and after L. (c) F1 Score: The median after L is greater than for S. The shape of the distribution is identical.

\begin{figure}[!t]
\centering
\subfloat{
\includegraphics[width=0.325\columnwidth]{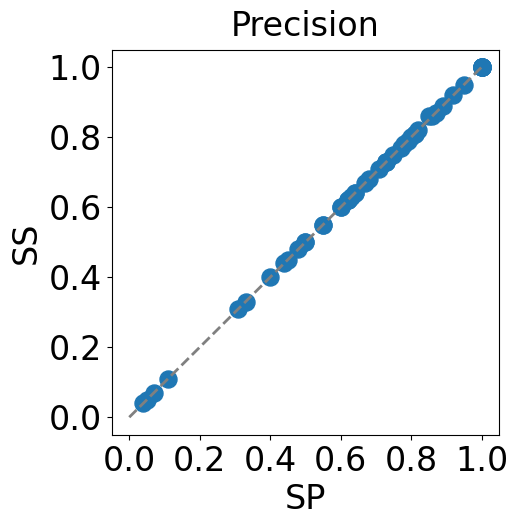}
\includegraphics[width=0.325\columnwidth]{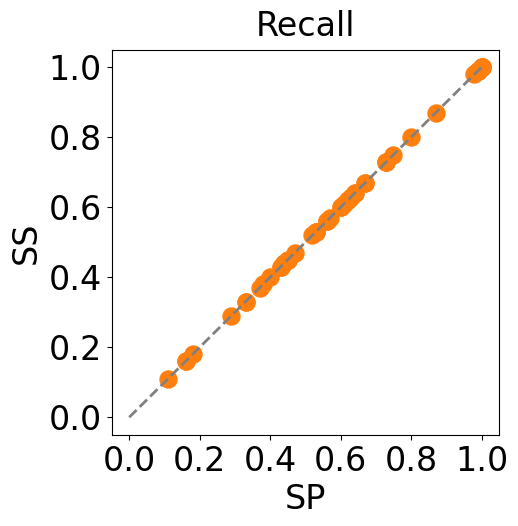}
\includegraphics[width=0.325\columnwidth]{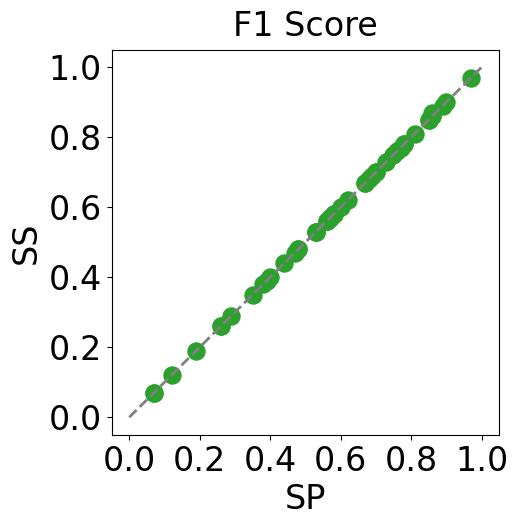}
}
\hfill
\subfloat{
\includegraphics[width=0.325\columnwidth]{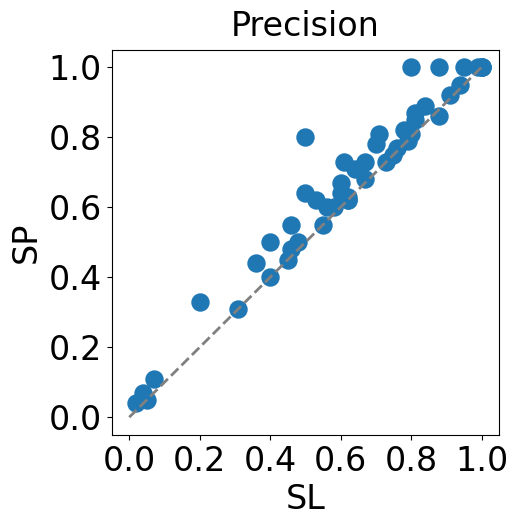}
\includegraphics[width=0.325\columnwidth]{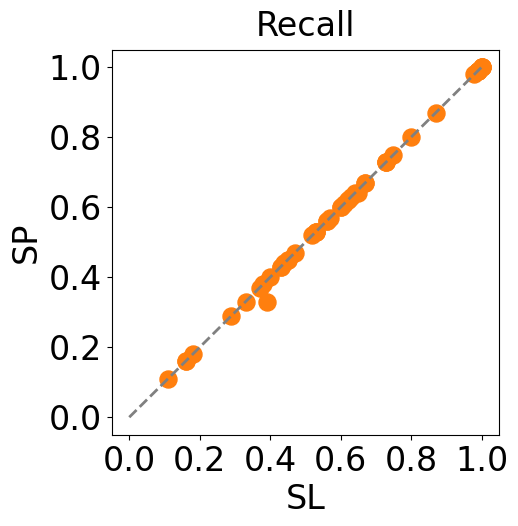}
\includegraphics[width=0.325\columnwidth]{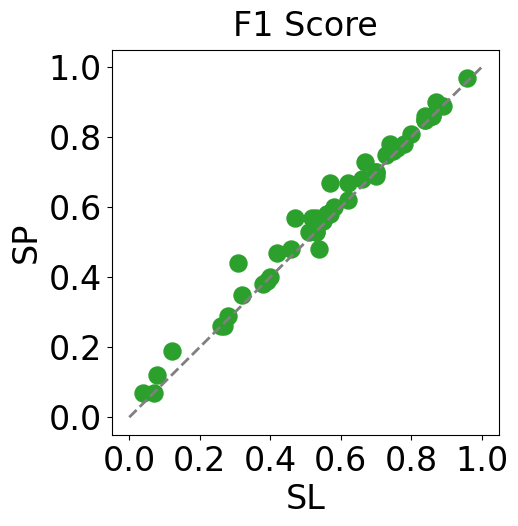}
}
\hfill
\subfloat{
\includegraphics[width=0.325\columnwidth]{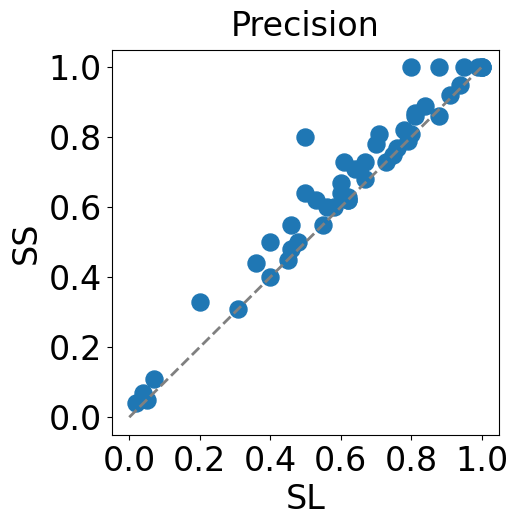}
\includegraphics[width=0.325\columnwidth]{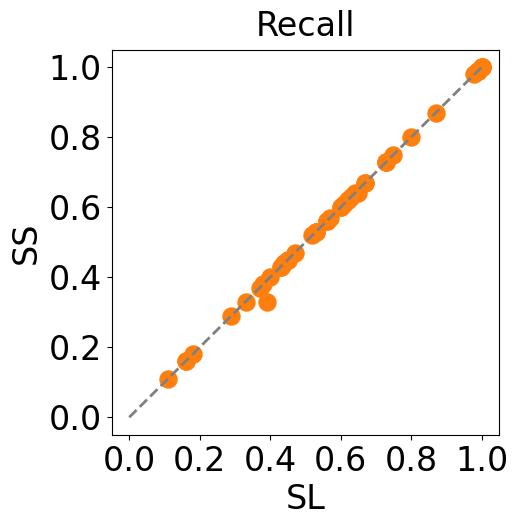}
\includegraphics[width=0.325\columnwidth]{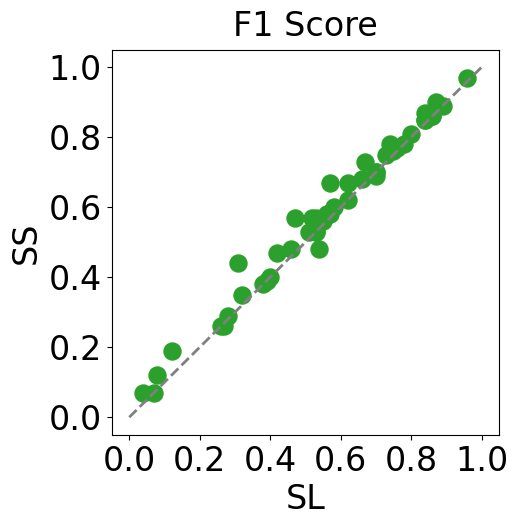}
}
\hfill
\subfloat{
\includegraphics[width=0.325\columnwidth]{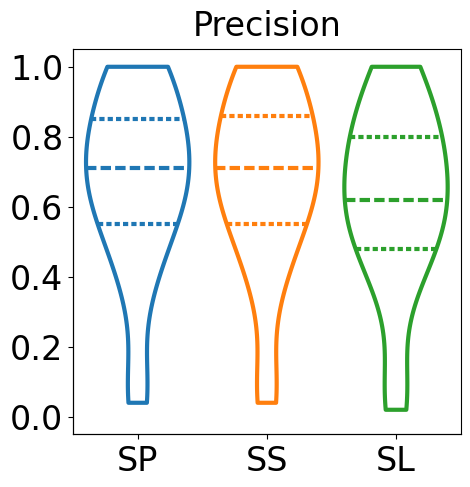}
\includegraphics[width=0.325\columnwidth]{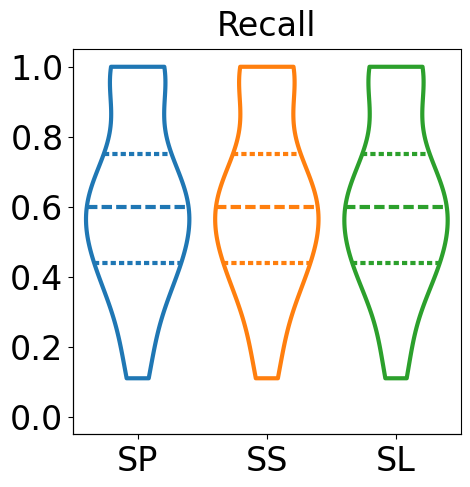}
\includegraphics[width=0.325\columnwidth]{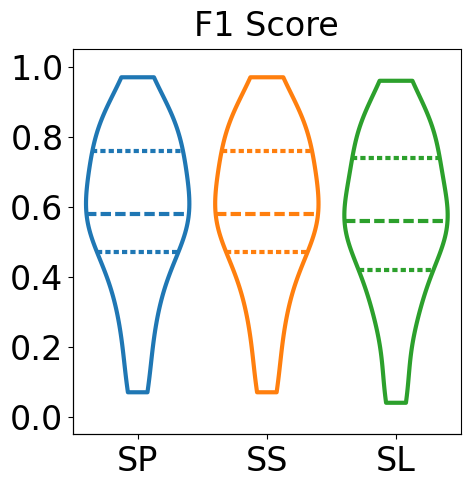}
}
\caption{Comparison of different stemming methods: Porter Stemmer (SP), Snowball Stemmer (SS), and Lancaster Stemmer (SL).}
\label{fig: compare-s}
\end{figure}

Fig.~\ref{fig: compare-s} compares different stemming methods: Porter Stemmer (SP), Snowball Stemmer (SS), and Lancaster Stemmer (SL). For SP and SS, all data points located on the equivalence line indicate that there is no difference between SP and SS. For SP/SS and SL, most data points above the equivalence line in precision and F1 score indicate that SP/SS is more effective than SL. In the violin plots: (a) Precision: The median number of SP and SS is greater than that of SL, and there is no difference between SP and SS. The shape of the distribution is identical. (b) Recall: The median number and the shape of the distribution are identical in SP, SS, and SL. (c) F1 Score: The median number of SP and SS is greater than that of SL, and there is no difference between SP and SS. The shape of the distribution is identical in SP, SS, and SL.

\begin{figure}[!t]
\centering
\subfloat{
\includegraphics[width=0.325\columnwidth]{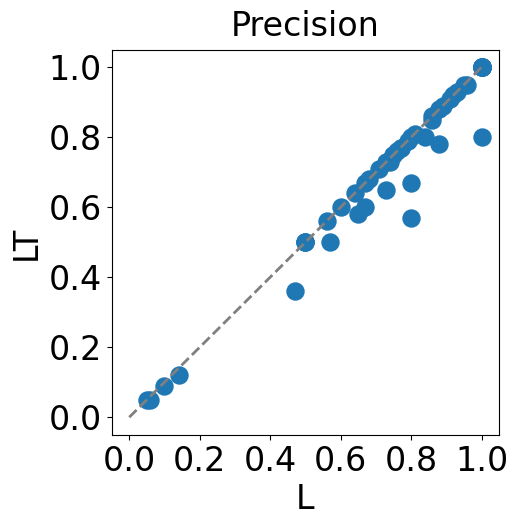}
\includegraphics[width=0.325\columnwidth]{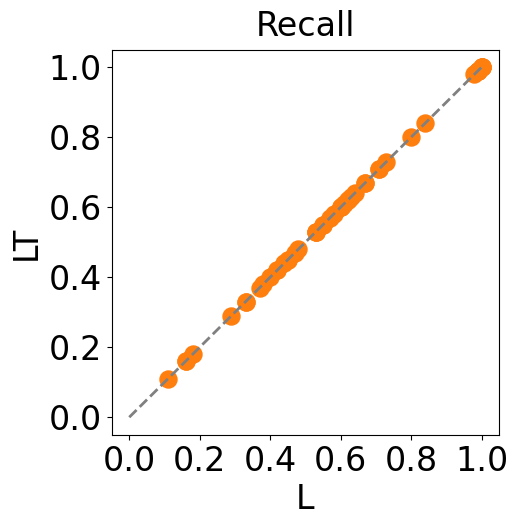}
\includegraphics[width=0.325\columnwidth]{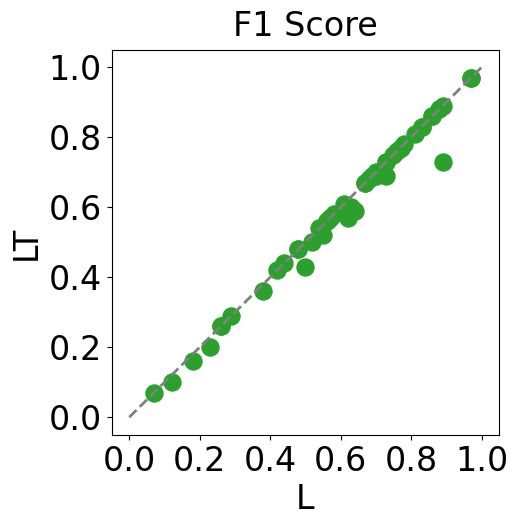}
}
\hfill
\subfloat{
\includegraphics[width=0.325\columnwidth]{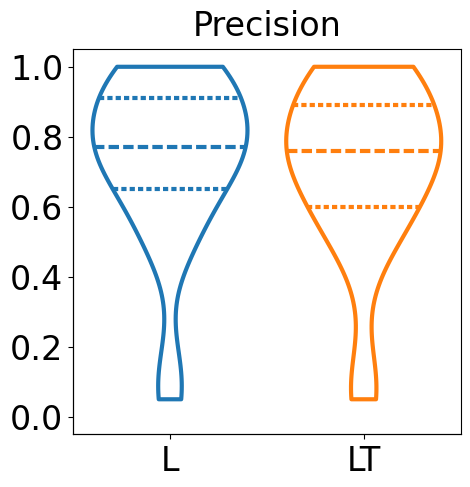}
\includegraphics[width=0.325\columnwidth]{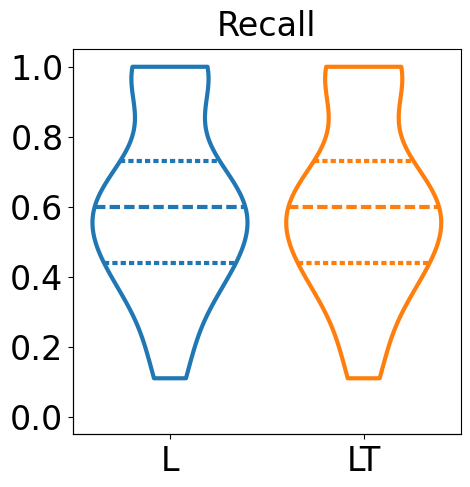}
\includegraphics[width=0.325\columnwidth]{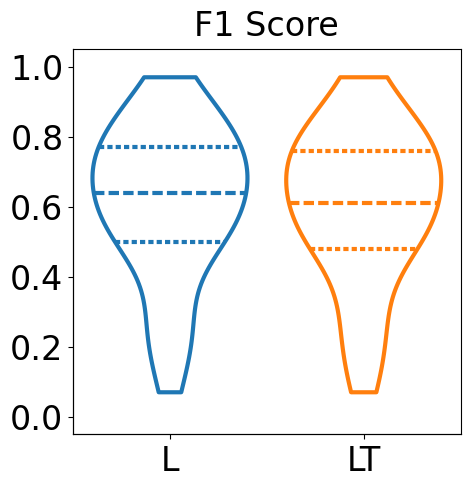}
}
\caption{Comparison of Lemmatisation (L) without/with POS Tagging (LT).}
\label{fig: compare-l}
\end{figure}

Fig.~\ref{fig: compare-l} compares Lemmatisation (L) and Lemmatisation + POS Tagging (LT). All data points located on the equivalence line indicate that using POS Tagging in Lemmatisation does not help OM. In the violin plots, for each of (a) Precision, (b) Recall, and (c) F1 Score, the median number and the shape of the distribution are identical for L and LT.

\subsection{Discussion}

For OM, the text preprocessing pipeline can be categorised into two phases. Phase 1 text preprocessing methods contain Normalisation and Tokenisation, whereas Phase 2 preprocessing methods consist of Stop Words Removal and Stemming/Lemmatisation. Phase 1 text preprocessing methods do not change word semantics, but change syntactic features such as formatting and typography. In contrast, Phase 2 text preprocessing methods change words to better capture semantic similarity, such as removing prefixes and suffixes or substituting common etymological roots.

For Phase 1 text preprocessing methods, we observe that matching performance usually increases with each method. In most cases, OM only requires minor preprocessing using Phase 1 text preprocessing methods. This is because entity names in the ontology are often compound words that do not occur in natural language, but they are partially formalised by agreement. There have been well-defined conventions established over the years, e.g. singularity, positive names, nouns for classes and verbs for properties~\cite{schober2007towards,odp2015}. However, Phase 1 text preprocessing methods could also benefit from customisation in OM tasks. Some traditional methods originating from natural language processing (NLP) are not applicable to OM and require adjustment. For example, sentence segmentation (i.e. splitting long text into sentences) is not applicable for ontology entities because they are generally short text fragments and do not need such operations. Word tokenisation (i.e. breaking text into single words) could be rewritten to detect abbreviations commonly used in ontology concept names and to tackle the conflicting use of Pascal case and snake case.

For Phase 2 text preprocessing methods, we observe no benefit in using these methods in OM. In some cases, Stop Words Removal and Stemming/Lemmatisation may even hamper the matching performance. There are plausible explanations for this behaviour arising from the nature of ontologies as distinct from natural language text, as follows:

\begin{enumerate}[wide, noitemsep, topsep=0pt, labelindent=0pt, label=(\alph*)]
\item Stop Words Removal (R): Logical expressions ``and'', ``or'', and ``not'' are stop words in English and usually carry little useful information, whereas these words express logical operations in ontology entities and may carry important semantics. Some single stop word characters (e.g. ``d'', ``i'', ``m'', ``o'', ``s'', ``t'', and ``y'') should be removed from the stop word set because contractions (e.g. I'd, I'm, o'clock, it's, isn't, y'know) are unlikely to appear in ontology entities, and yet those single characters are likely to carry important meaning if they do occur (e.g. ``Chromosome\_Y'').
\item Stemming (S) vs Lemmatisation (L): While L tends to avoid generating FPs, it may also miss some implicit TPs. On the other hand, S is more eager to find TPs, but eagerness can also lead to more FPs being found. Based on our experiments, L is superior to S without post hoc correction.
\item Different stemming methods: Porter Stemmer (PS) and Snowball Stemmer (SS) have been found to perform better than Lancaster Stemmer (LS), and there is no significant difference between PS and SS. Although LS is more eager to detect more TPs, it does not lead to performance improvement as more FPs are also matched.
\item Lemmatisation (L) vs Lemmatisation + POS Tagging (LT): LT is generally expected to have better results than L alone because tagging can help detect more precise root words. However, we do not observe such a performance improvement when using POS tagging in our study. The reason may be that ontology classes are usually nouns or gerunds, and in such cases, we could expect the simpler grammatical assumption to have similar results.
\end{enumerate}

\section{Text Preprocessing Pipeline Repair}
\label{sec: repair}

The experimental results in Section~\ref{sec: analysis} demonstrate that only Phase 1 text preprocessing (Tokenisation and Normalisation) helps with OM. The use of Phase 2 text preprocessing (Stop Words Removal and Stemming/Lemmatisation) does not improve performance and may even have negative impacts. In this section, we explore the underlying cause of this behaviour and propose our novel pipeline repair approaches.

\subsection{Causal Analysis}

Phase 1 text preprocessing methods (Tokenisation and Normalisation) do not change the text meaning. For example, {\itshape isReviewing} is equivalent to {\itshape is\_reviewing}. This means that applying Phase 1 methods only helps to detect TPs, while the number of FPs remains unchanged. For this reason, Phase 1 methods always have a positive effect on precision, recall, and overall F1 score (Equations~\ref{equation: precision-1},~\ref{equation: recall-1}, and~\ref{equation: f1-1}).

\begin{equation}
\label{equation: precision-1}
Prec \uparrow  = \frac{|A \cap R|}{|A|} = \frac{TP}{TP+FP} = 1 - \frac{FP}{TP \uparrow + FP}
\end{equation}

\begin{equation}
\label{equation: recall-1}
Rec \uparrow \ = \frac{|A \cap R|}{|R|} = \frac{TP\uparrow}{|R|}
\end{equation}

\begin{equation}
\label{equation: f1-1}
F_1 \ Score \uparrow = \frac{2}{Prec \uparrow ^{-1} + Rec\uparrow ^{-1}}
\end{equation}

Phase 2 text preprocessing methods (Stop Words Removal and Stemming/Lemmatisation) are actually relaxations of matching rules in OM tasks. Moving through the text preprocessing pipeline tends to detect more TPs and FPs in the derived alignment (A). For example, {\itshape isReviewing} and {\itshape isReviewedBy} may be object properties with distinctly different meanings, but removing the common stop words ``is'' and ``by'', and using Stemming/Lemmatisation to retrieve the same root word ``Review'', could cause a false match. For this reason, recall is always increasing (Equation~\ref{equation: recall-2}), but precision and the overall F1 score are less reliable (Equations~\ref{equation: precision-2} and~\ref{equation: f1-2}).

\begin{equation}
\label{equation: precision-2}
Prec \ ? = \frac{|A \cap R|}{|A|} = \frac{TP\uparrow}{TP\uparrow + FP\uparrow}
\end{equation}

\begin{equation}
\label{equation: recall-2}
Rec \uparrow \ = \frac{|A \cap R|}{|R|} = \frac{TP\uparrow}{|R|}
\end{equation}

\begin{equation}
\label{equation: f1-2}
F_1 \ Score \ ? = \frac{2}{Prec \ ? ^{-1} + Rec\uparrow ^{-1}}
\end{equation}

If we define $\Delta TP$ and $\Delta FP$ as the increase in TP and FP for a preprocessing method, then Equation~\ref{equation: condition} presents the condition to increase precision and F1 score.

\begin{equation}
\label{equation: condition}
\frac{TP+\Delta TP}{TP+\Delta TP+FP+\Delta FP} > \frac{TP}{TP+FP} \Rightarrow \frac{\Delta TP}{\Delta FP} > \frac{TP}{FP}
\end{equation}

In our experiments, we actually observe a reduction in precision and overall F1 score. This means that Phase 2 methods produce more FPs than TPs, and this proportion is less than the original number of $TP/FP$. So performance does not improve, unless the benefit of each TP is considered more valuable than the disbenefit of each FP. This could apply, for example, if we are expecting a post hoc correction phase in which removing FPs is considered to be an easier human task than adding missing TPs. This idea inspires our novel pipeline repair method.

\subsection{Motivation}

\begin{figure}[htbp]
\centering
\includegraphics[width=1\columnwidth]{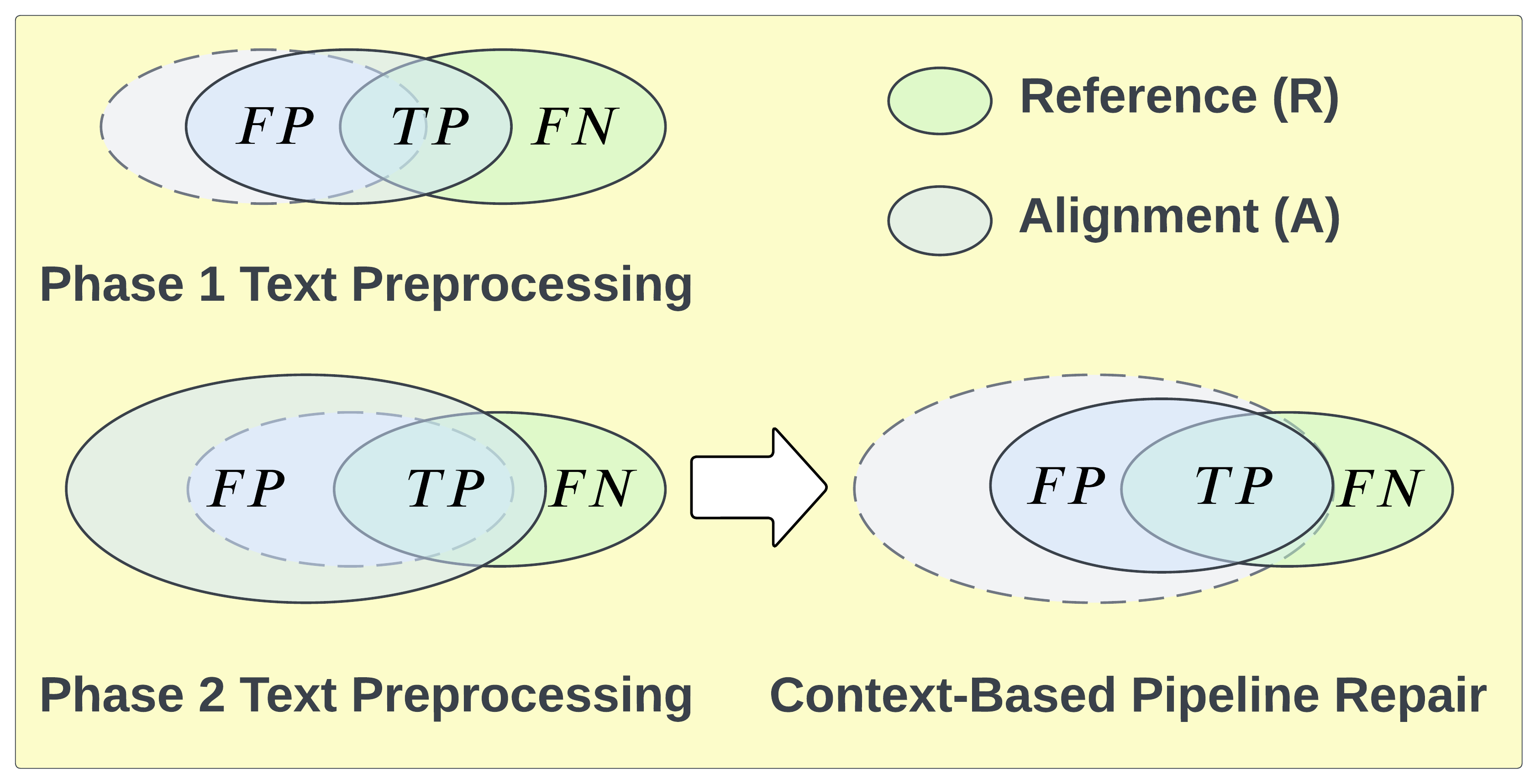}
\caption{Motivation of our context-based pipeline repair.}
\label{fig: summary}
\end{figure}

While Phase 2 text preprocessing methods have been shown to be less effective in OM by introducing FPs, we propose a pipeline repair approach to differentiate FPs from TPs. As illustrated in Fig.~\ref{fig: summary}: (1) Phase 1 methods \textbf{shift Alignment (A) towards Reference (R)}. The number of TPs increases, while the number of FPs decreases. (2) Phase 2 methods \textbf{expand Alignment (A)}. The number of TPs increases, but the number of FPs also increases. (3) Our text preprocessing pipeline repair aims to \textbf{collapse Alignment (A)}. The number of FPs decreases significantly, with a slight decrease in TPs.

\subsection{Pre hoc Logic-Based Repair}

The pre hoc logic-based repair is applied before text preprocessing. One critical step in our approach is to assemble a set of \emph{reserved words} that may cause FPs in text preprocessing, and to exclude these words from text preprocessing. The selection criteria are based on two ontology features: (1) entity names (class or property) are not duplicated within a single ontology; and (2) ontologies that represent the same domain knowledge tend to use similar terminologies (see Section~\ref{sec: background and related work}). This approach is context-based because pairs of words may have the same or different meanings in different contexts. Based on these two features, we propose a simple Algorithm~\ref{alg: repair} to assemble the reserved word set. For both $O_s$ and $O_t$, the algorithm first iterates on all pairs of entities $e_i, e_j $ in the ontology, respectively. For a specific text preprocessing method $f(\cdot)$, if $f(e_i) = f(e_j)$, we retrieve the different words between $e_i$ and $e_j$ and store them in the reserved word set. We define $S_i = \{i_1,..,p,..,i_n\} $ is the string set of $e_i$ and $S_j = \{j_1,...,p,...j_m\} $ is the string set of $e_j$. Let $p$ be the longest common substrings in both $S_i$ and $S_j$, then $S_e \div S_j = \{i_1,..,i_n,j_1,...,j_m\}$. If a word appears in the reserved set, the text preprocessing pipeline skips the operation for this word. To simplify the reserved word set, we remove the words where $f(w) = w$ because their occurrence in the reserved word set has no effect on mapping results.

\begin{algorithm}[htbp]
\caption{Finding Reserved Word Set}
\label{alg: repair}
\begin{algorithmic}
\renewcommand{\algorithmicrequire}{\textbf{Input:}}
\renewcommand{\algorithmicensure}{\textbf{Output:}}
\REQUIRE Source Ontology $O_s$, Target Ontology $O_t$, \\
Text Preprocessing Pipeline Method $f(\cdot)$
\ENSURE Reserved\_Word\_Set $S_r$\\

\STATE \textbf{/* Phase 1: Find duplicates in $O_s$ and $O_t$ */}\\

\STATE {/* Find duplicates in $O_s$ */}\\

\FOR {$ Entity \ e_i, e_j \in O_s $}
\IF {$ f(e_i)=f(e_j)$}
\STATE {$ S_r = S_r \cup (S_i \div S_j) $}
\ENDIF
\ENDFOR

\STATE {/* Find duplicates in $O_t$ */}\\
\STATE {/* Same procedure applies ...*/}

\STATE \textbf{/* Phase 2: Find immutable words in $S$ */}\\
\FOR {$ Word \ w \in S $}
\IF {$ f(w)=w $}
\STATE {$ S_r = S_r \setminus \{w\} $}
\ENDIF
\ENDFOR

\RETURN Reserved\_Word\_Set $S_r$
\end{algorithmic}
\end{algorithm}

For example, consider two object properties \textit{was\_a\_member\_of} and \textit{has\_members} from a single ontology that have the same result ``member'' via traditional text preprocessing methods. Because there are no duplicate entities within a single ontology, we use a reserved word set in our proposed pipeline repair approach to determine that they are distinguishing entities. The initial step (i.e. Phase 1 of Algorithm~\ref{alg: repair}) is to add [``was'', ``a'', ``member'', ``of'', ``has'', ``members''] to the reserved word set so that these two object properties would not have the same text preprocessing results. \textit{was\_a\_member\_of} preprocessed with skipping the reserved word set is ``was a member of'', while \textit{has\_members} preprocessed with the reserved word set is ``has members''. The revision step (i.e. Phase 2 of Algorithm~\ref{alg: repair}) is to check whether there are immutable words in the reserved set. We can observe that the word ``member'' is immutable because it is the same before and after text preprocessing. Removing this word from the reserved word set still makes the two object properties different. Therefore, the final reserved word set is [``was'', ``a'', ``of'', ``has'', ``members''].

The generated reserved word set can be used to repair false mappings between entities within the same domain context but coming from different ontologies. For example, we expect that the two object properties \textit{has\_a\_steering\_committee} and \textit{was\_a\_steering\_committee\_of} are non-identical. While a false mapping may occur when they both have the same result ``steer committe'' after the traditional text preprocessing, using the reserved word set can repair this false mapping. As the words ``has'', ``a'', ``was'', and ``of'' are listed as reserved words, these two named properties are processed as ``has a steer committe'' and ``was a steer committe of'', respectively.

\begin{figure}[!ht]
\centering
\hfill
\subfloat{
\includegraphics[width=0.325\columnwidth]{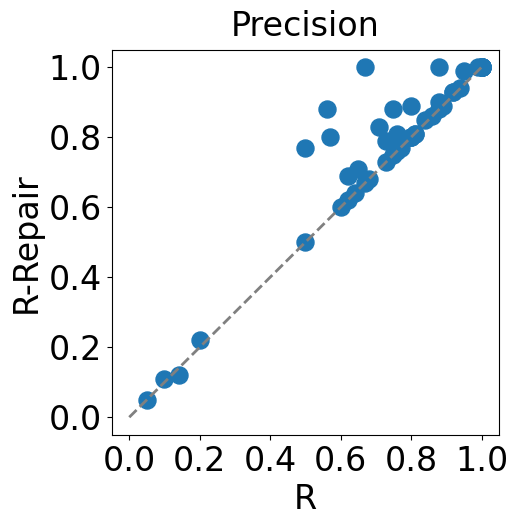}
\includegraphics[width=0.325\columnwidth]{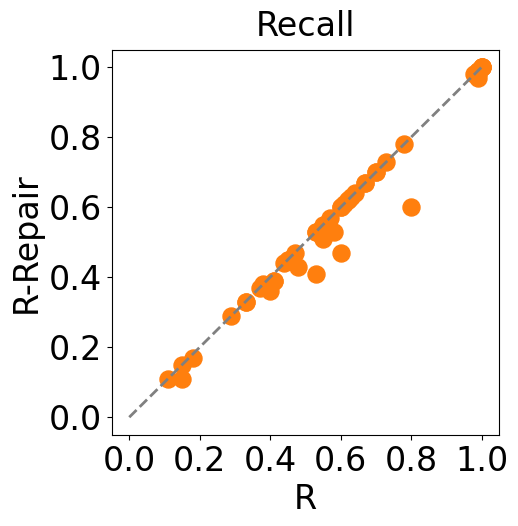}
\includegraphics[width=0.325\columnwidth]{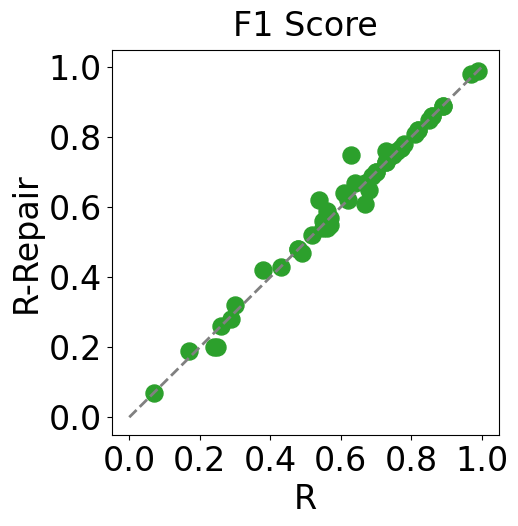}
}
\hfill
\subfloat{
\includegraphics[width=0.325\columnwidth]{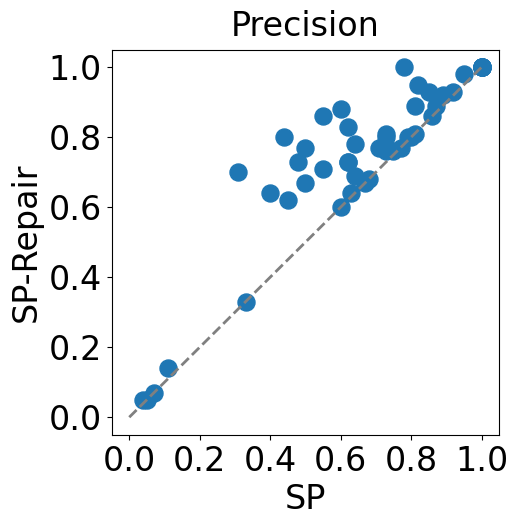}
\includegraphics[width=0.325\columnwidth]{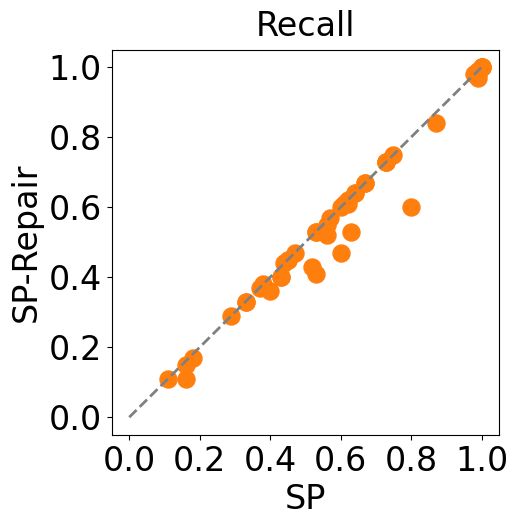}
\includegraphics[width=0.325\columnwidth]{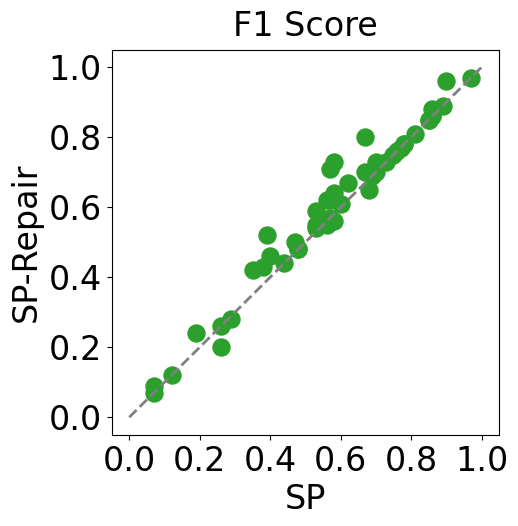}
}
\hfill
\subfloat{
\includegraphics[width=0.325\columnwidth]{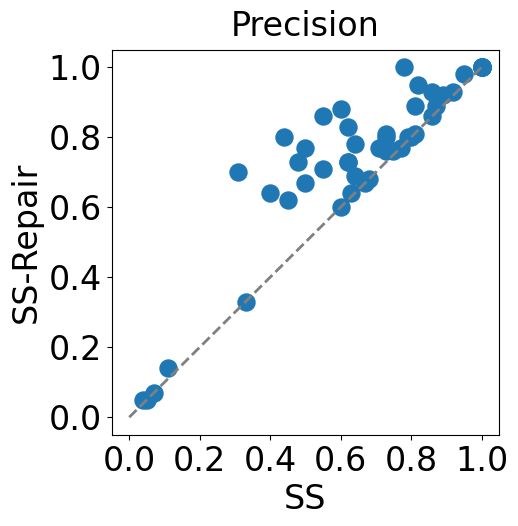}
\includegraphics[width=0.325\columnwidth]{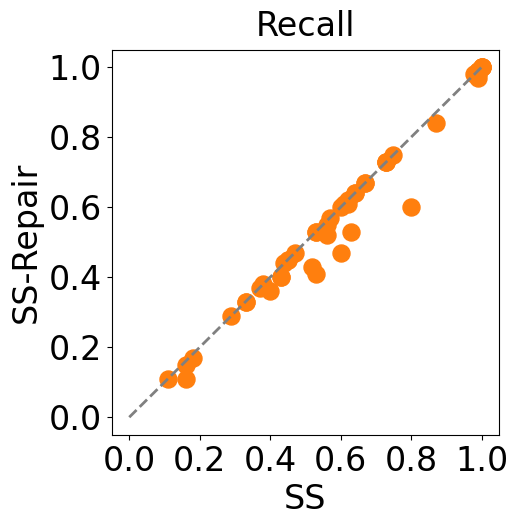}
\includegraphics[width=0.325\columnwidth]{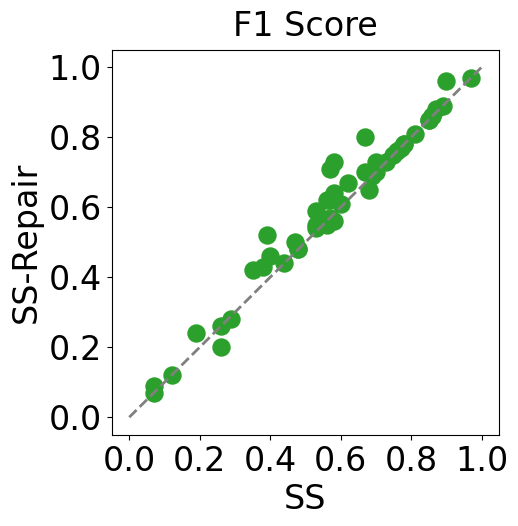}
}
\hfill
\subfloat{
\includegraphics[width=0.325\columnwidth]{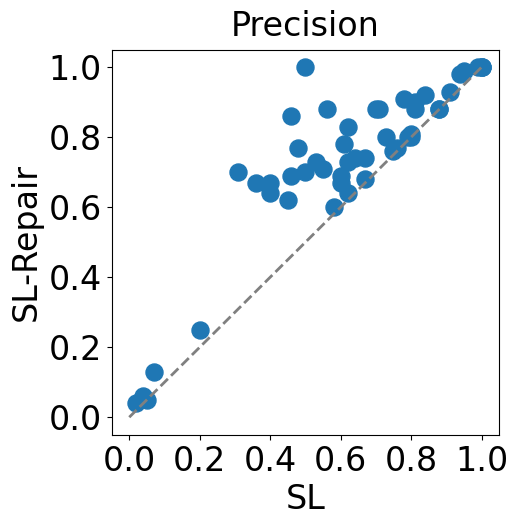}
\includegraphics[width=0.325\columnwidth]{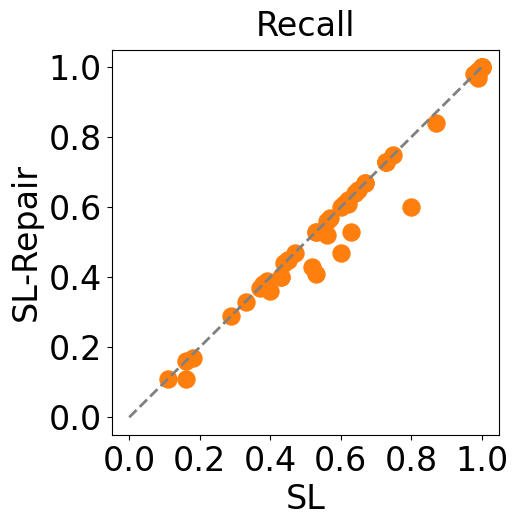}
\includegraphics[width=0.325\columnwidth]{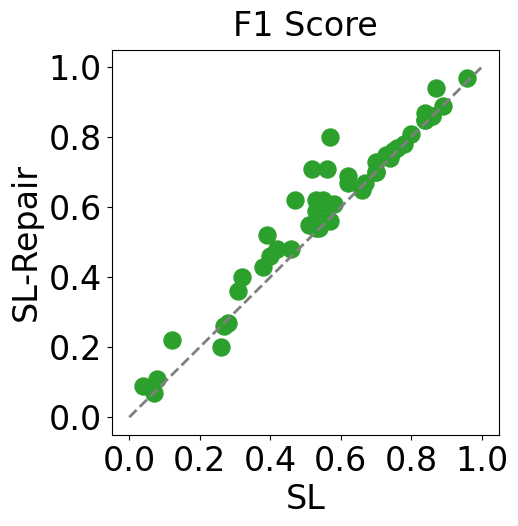}
}
\hfill
\subfloat{
\includegraphics[width=0.325\columnwidth]{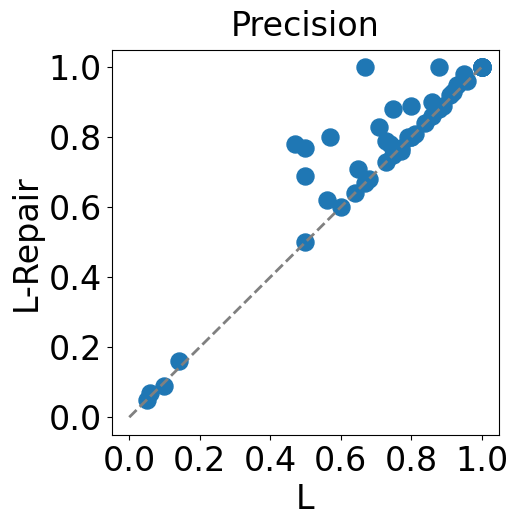}
\includegraphics[width=0.325\columnwidth]{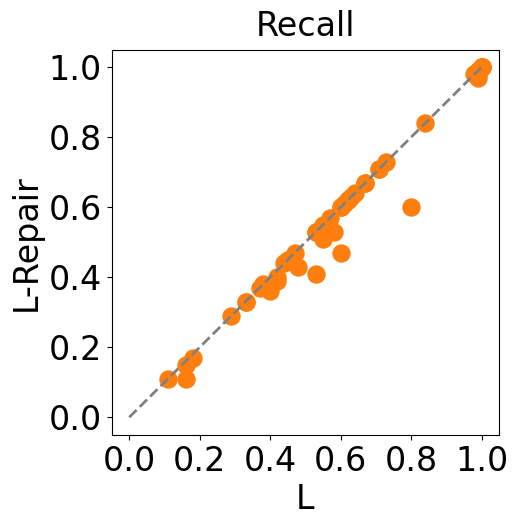}
\includegraphics[width=0.325\columnwidth]{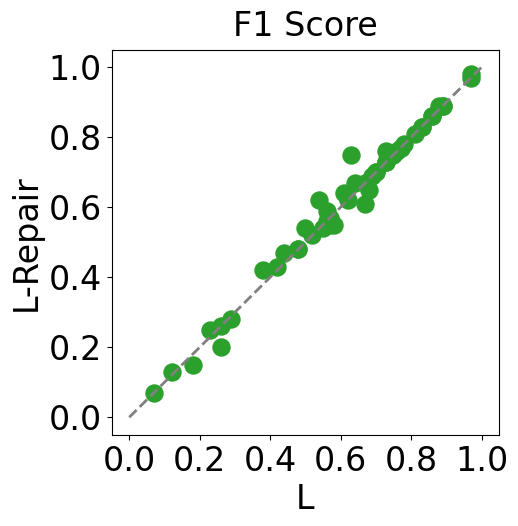}
}
\hfill
\subfloat{
\includegraphics[width=0.325\columnwidth]{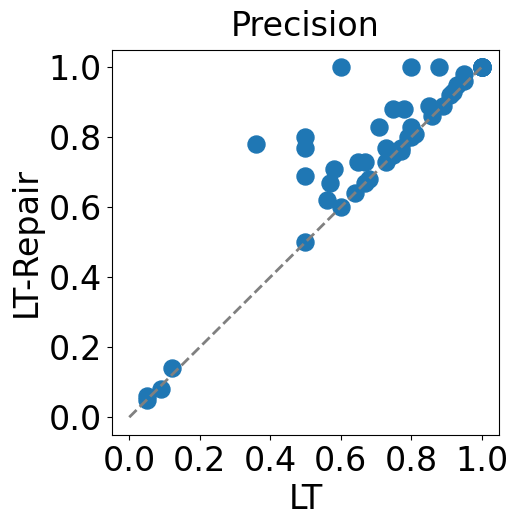}
\includegraphics[width=0.325\columnwidth]{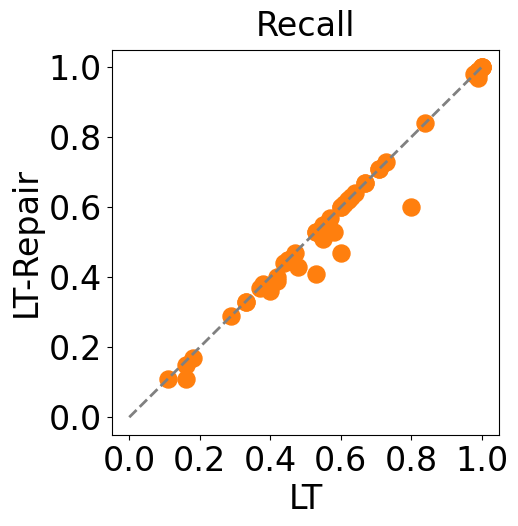}
\includegraphics[width=0.325\columnwidth]{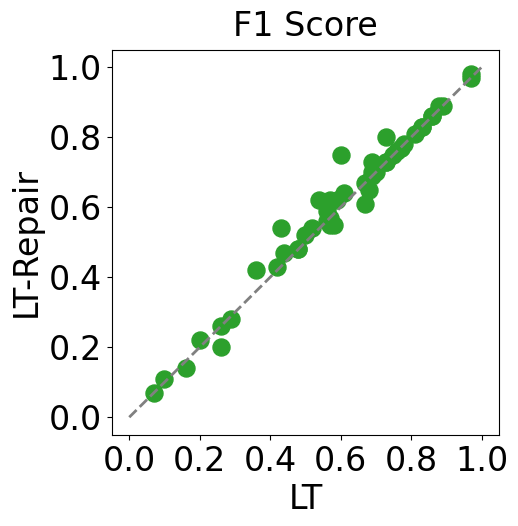}
}
\caption{Context-based pipeline repair use in Phase 2 text
preprocessing: Stop Words Removal (R), Porter Stemmer (SP), Snowball Stemmer (SS), Lancaster Stemmer (SL), Lemmatisation (L), and Lemmatisation + POS Tagging (LT).}
\label{fig: context-based-repair}
\end{figure}

We apply our context-based pipeline repair approach to the same OAEI datasets as above. Fig.~\ref{fig: context-based-repair} compares with and without context-based pipeline repair in 8 tracks with 49 distinct alignments. The text preprocessing pipeline methods implemented include less effective Phase 2 text preprocessing methods: Stop Words Removal (R), Porter Stemmer (SP), Snowball Stemmer (SS), Lancaster Stemmer (SL), Lemmatisation (L), and Lemmatisation + POS Tagging (LT). For precision, most data points are above the equivalence line, indicating that context-based pipeline repair significantly improves matching correctness. For recall, most data points are located on the equivalence line, and only a few data points are below the equivalence line, indicating that context-based pipeline repair slightly reduces matching completeness. Looking at F1 score, most of the data points are above the equivalence line, indicating that context-based pipeline repair also improves overall matching performance.

Fig.~\ref{fig: context-based-repair} also shows the relative improvement in matching performance made by Phase 2 preprocessing methods. This can be seen from the distance of data points from the equivalence line. We see that the matching performance improvement has Stemming (S)~\textgreater~Lemmatisation (L)~\textgreater~Stop Words Removal (R). Further, the matching performance improvement with different stemmers has Lancaster Stemmer (SL)~\textgreater~Porter Stemmer (SP) = Snowball Stemmer (SS) and the matching performance improvement for lemmatisation  (L) alone is very similar to Lemmatisation + POS Tagging (LT).

\begin{figure}[htbp]
\centering
\includegraphics[width=0.75\columnwidth]{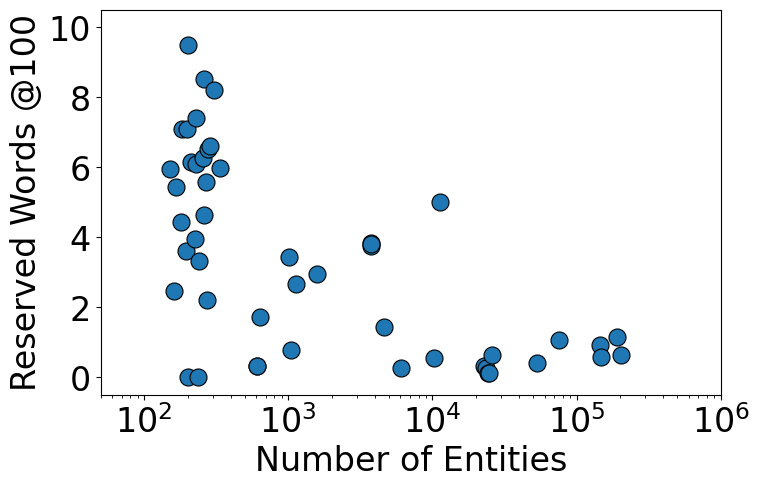}
\caption{Reserved words per 100 entities compared to the number of entities.}
\label{fig: ar-evaluation}
\end{figure}

Fig.~\ref{fig: ar-evaluation} evaluates the scalability of pre hoc logic-based repair. The x-axis indicates the number of entities in each alignment, and the y-axis shows the number of reserved words per 100 entities. For all the alignments we tested, the value of reserved words per 100 entities is less than 10. With the number of entities increasing (i.e. the ontology size increases), the number of reserved words per 100 entities decreases. This may be because reserved words such as``has'', ``is'', and ``a'' are commonly used in many ontologies and in multiple distinct entities within those ontologies. Conventionally, there are a limited number of such words, and therefore the reserved word feature of our approach is scalable in large-scale OM.

\subsection{Post Hoc LLM-Based Repair}
\label{sub-sec: llm-based-repair}

In this study, we aim to develop a generic pipeline repair approach and establish best practices for integrating LLMs in the repair phase by empirical analysis with selected OAEI tracks. LLMs vary with respect to prompting strategies that could be used in LLM-based repair.

We choose 10 LLMs from 4 different families. These include two OpenAI GPT models~\cite{llm-gpt} (gpt-4o and gpt-4o-mini), two Anthropic Claude models~\cite{llm-claude} (claude-3-sonnet and claude-3-haiku), and two versions of Meta Llama~\cite{llm-llama} (llama-3-8b and llama-3.1-8b), two versions of Alibaba Qwen~\cite{llm-qwen} (qwen-2-7b and qwen-2.5-7b), Google Gemma 2~\cite{llm-gemma}, and ChatGLM 4~\cite{glm2024chatglm}. We use LangChain~\cite{langchain} to build chats with OpenAI GPT and Anthropic Claude models, and Ollama~\cite{ollama} to access open-source models. Table~\ref{tab: llm-version} lists the details of selected LLMs (retrieved from~\cite{qiang2023agent}). All model temperatures are set to 0 to minimise the random results generated by LLMs.

\begin{table}[htbp]
\centering
\renewcommand\arraystretch{1.2}
\tabcolsep=0.15cm
\caption{Details of selected LLMs. API-accessed models are shown as triangles ($\blacktriangle$) and open-source models as circles ($\bullet$).}
\label{tab: llm-version}
\begin{adjustbox}{width=1\columnwidth,center}
\begin{tabular}{|c|l|l|} 
\hline
\multicolumn{1}{|c|}{\textbf{Family}}   & \multicolumn{1}{|l|}{\textbf{Model}}  & \multicolumn{1}{|l|}{\textbf{Version}}    \\ \hline
\multirow{2}{*}{GPT}                    & $\blacktriangle$ gpt-4o               & gpt-4o-2024-05-13                         \\ \cline{2-3}
                                        & $\blacktriangle$ gpt-4o-mini          & gpt-4o-mini-2024-07-18                    \\ \cline{1-3}
\multirow{2}{*}{Claude}                 & $\blacktriangle$ claude-3-sonnet      & claude-3-sonnet-20240229                  \\ \cline{2-3}
                                        & $\blacktriangle$ claude-3-haiku       & claude-3-haiku-20240307                   \\ \cline{1-3}
\multirow{2}{*}{Llama}                  & \filledcirc~llama-3-8b                & Ollama Model ID: 365c0bd3c000             \\ \cline{2-3}
                                        & \filledcirc~llama-3.1-8b              & Ollama Model ID: 46e0c10c039e             \\ \cline{1-3}
\multirow{2}{*}{Qwen}                   & \filledcirc~qwen-2-7b                 & Ollama Model ID: dd314f039b9d             \\ \cline{2-3}
                                        & \filledcirc~qwen-2.5-7b               & Ollama Model ID: 845dbda0ea48             \\ \cline{1-3}       
\multirow{1}{*}{Gemma}                  & \filledcirc~gemma-2-9b                & Ollama Model ID: ff02c3702f32             \\ \cline{1-3}
\multirow{1}{*}{GLM}                    & \filledcirc~glm-4-9b                  & Ollama Model ID: 5b699761eca5             \\ \hline
\end{tabular}
\end{adjustbox}
\end{table}

We select 4 different LLM prompt strategies. (a) Basic: The simplest prompt template to compare two entities using ``Is Entity1 equivalent to Entity2? Answer yes or no.'' (b) Few-Shot: Adding an example in the prompt. We use the example ``Example: Hair\_root is equivalent to Hair\_Root.'' (c) Self-Reflection: Before LLMs answer the question, ask LLMs to rethink the output and give a short explanation. We add a sentence ``Write a short explanation.'' to invoke this function. (d) Few-Shot + Self-Reflection: A combination of (b) and (c). Table~\ref{tab: prompt-templates} lists the details of the selected prompt strategies and their templates.

\begin{table}[!ht]
\renewcommand\arraystretch{1.2}
\tabcolsep=0.15cm
\caption{Selected prompt templates. The few-shot parts are coloured blue, and the self-reflection parts are orange.}
\label{tab: prompt-templates}
\begin{adjustbox}{width=1\columnwidth,center}
\begin{tabular}{|c|c|p{0.8\columnwidth}|}
\hline
\multicolumn{1}{|c}{\textbf{No.}}   & \multicolumn{1}{|c|}{\textbf{Prompt Strategy}}   & \multicolumn{1}{|l|}{\textbf{Prompt Template}}     \\ \hline
\multirow{1}{*}{PT1}                & \multirow{1}{*}{Zero-Shot}                & Is Entity1 equivalent to Entity2? Answer yes or no.       \\ \hline
\multirow{2}{*}{PT2}                & \multirow{2}{*}{Few-Shot}                 & \textcolor{blue}{Example: Hair\_root is equivalent to Hair\_Root.}    \\ 
                                    &                                           & Is Entity1 equivalent to Entity2? Answer yes or no.       \\ \hline
\multirow{2}{*}{PT3}                & \multirow{2}{*}{Self-Reflection}          & Is Entity1 equivalent to Entity2? Answer yes or no.       \\
                                    &                                           & \textcolor{orange}{Write a short explanation.}            \\ \hline
\multirow{3}{*}{PT4}                & \multirow{1}{*}{Few-Shot}                 & \textcolor{blue}{Example: Hair\_root is equivalent to Hair\_Root.}    \\
                                    & \multirow{1}{*}{+}                        & Is Entity1 equivalent to Entity2? Answer yes or no.                   \\
                                    & \multirow{1}{*}{Self-Reflection}          & \textcolor{orange}{Write a short explanation.}            \\ \hline
\end{tabular}
\end{adjustbox}
\end{table}

We experiment with the \textit{Largebio} Track that contains pairwise alignments between FMA, NCI, and SNOMED. We choose three datasets: WHOLE-FMA-NCI, WHOLE-FMA-SNOMED, and WHOLE-SNOMED-NCI. We compare LLMs with the classical text preprocessing pipeline in detecting TPs and FPs, which are two main contributors to precision and recall in matching performance. The true mappings (TPs) are retrieved from the reference (R) where the mappings can be found by OM. The false mappings (FPs) are retrieved from the alignment (A) generated by the classical text preprocessing pipeline but exclude TPs above. In this study, we report our results on a single run. There may be slight differences in the results across multiple trials, but these differences are not significant in terms of the ranking and trend of different LLMs and prompt templates.

\begin{figure*}[htbp]
\centering
\subfloat[PT1: WHOLE-FMA-NCI]{\includegraphics[width=0.33\textwidth]{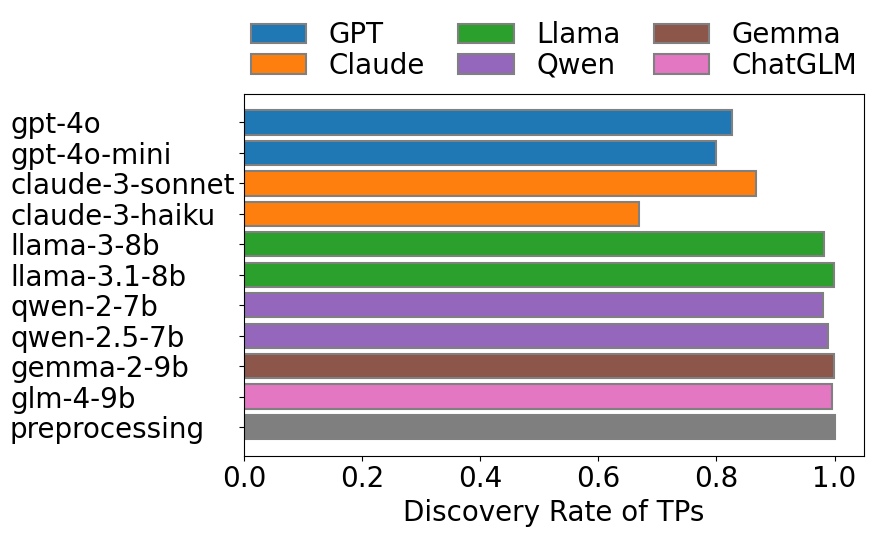}}
\subfloat[PT1: WHOLE-FMA-SNOMED]{\includegraphics[width=0.33\textwidth]{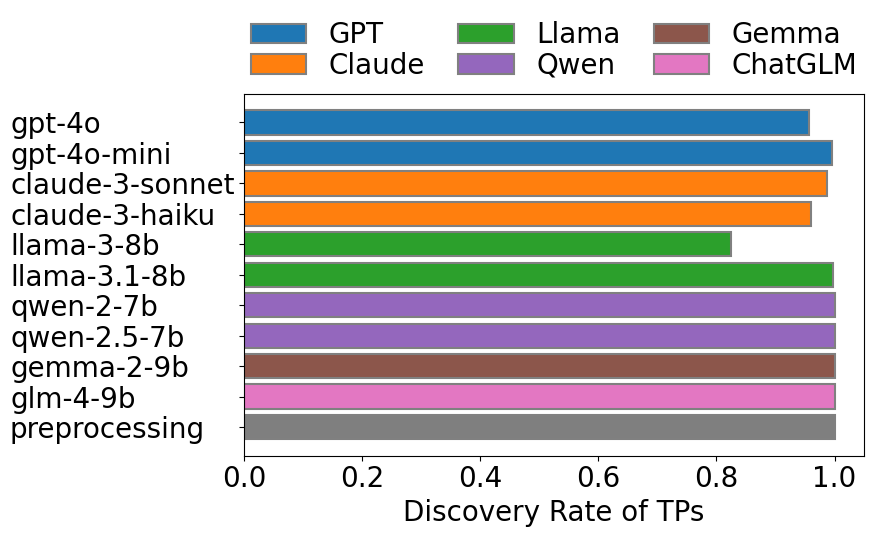}}
\subfloat[PT1: WHOLE-SNOMED-NCI]{\includegraphics[width=0.33\textwidth]{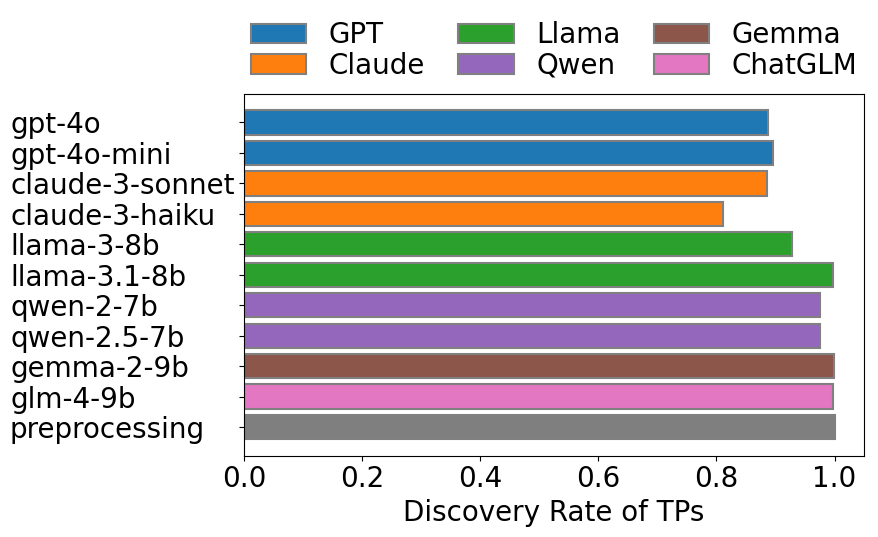}}                          \\
\subfloat[PT2: WHOLE-FMA-NCI]{\includegraphics[width=0.33\textwidth]{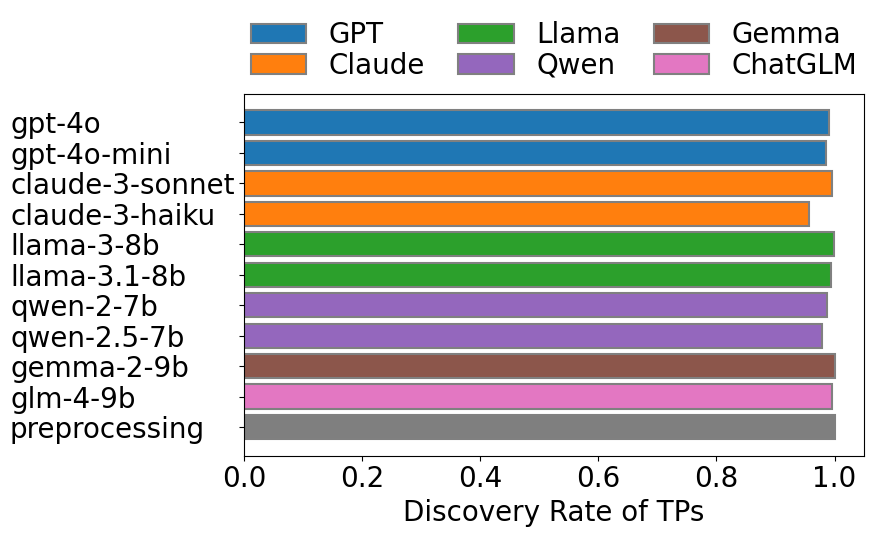}}
\subfloat[PT2: WHOLE-FMA-SNOMED]{\includegraphics[width=0.33\textwidth]{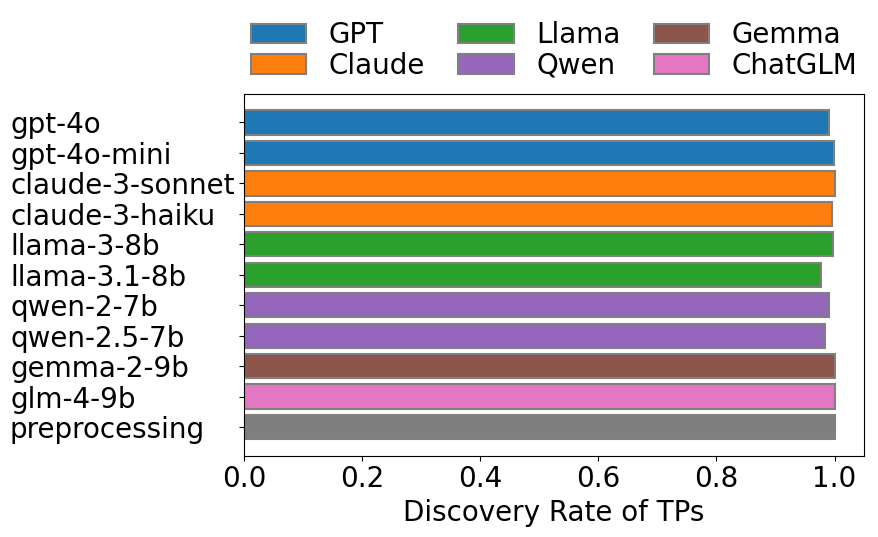}}
\subfloat[PT2: WHOLE-SNOMED-NCI]{\includegraphics[width=0.33\textwidth]{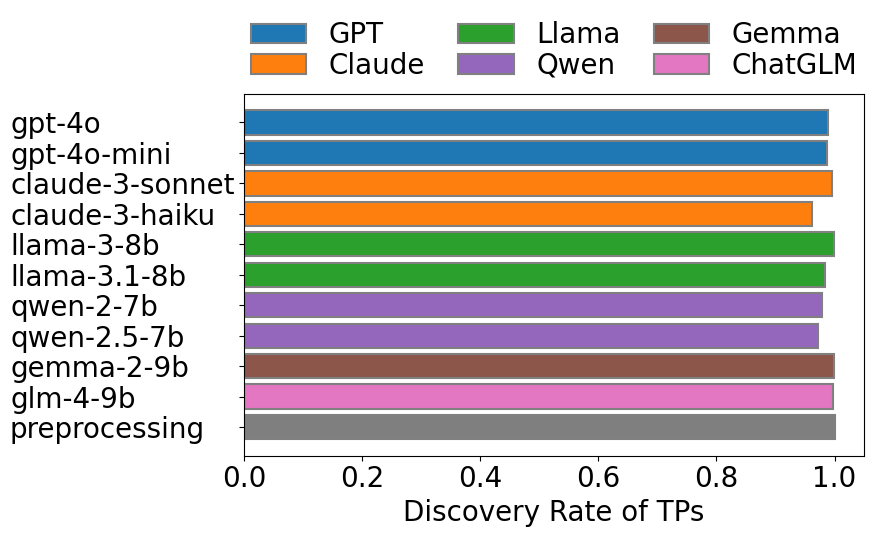}}                       \\
\subfloat[PT3: WHOLE-FMA-NCI]{\includegraphics[width=0.33\textwidth]{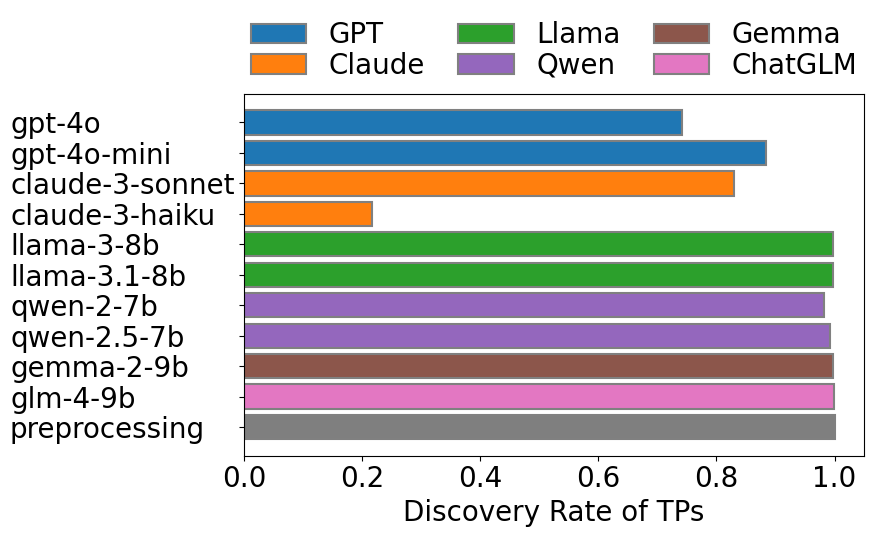}}
\subfloat[PT3: WHOLE-FMA-SNOMED]{\includegraphics[width=0.33\textwidth]{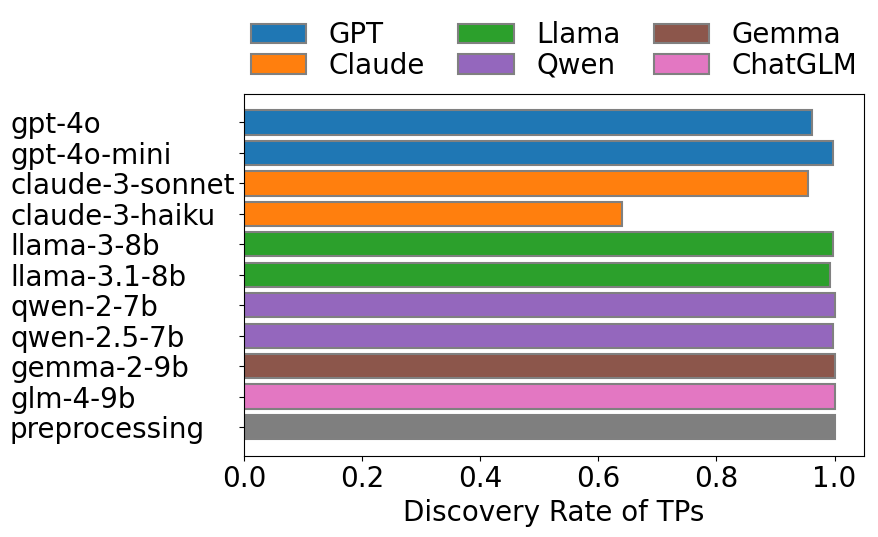}}
\subfloat[PT3: WHOLE-SNOMED-NCI]{\includegraphics[width=0.33\textwidth]{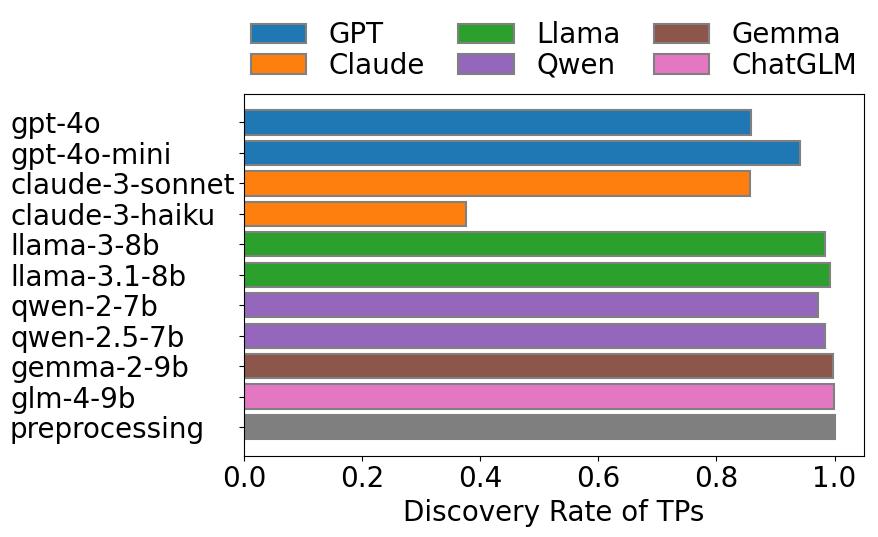}}                \\
\subfloat[PT4: WHOLE-FMA-NCI]{\includegraphics[width=0.33\textwidth]{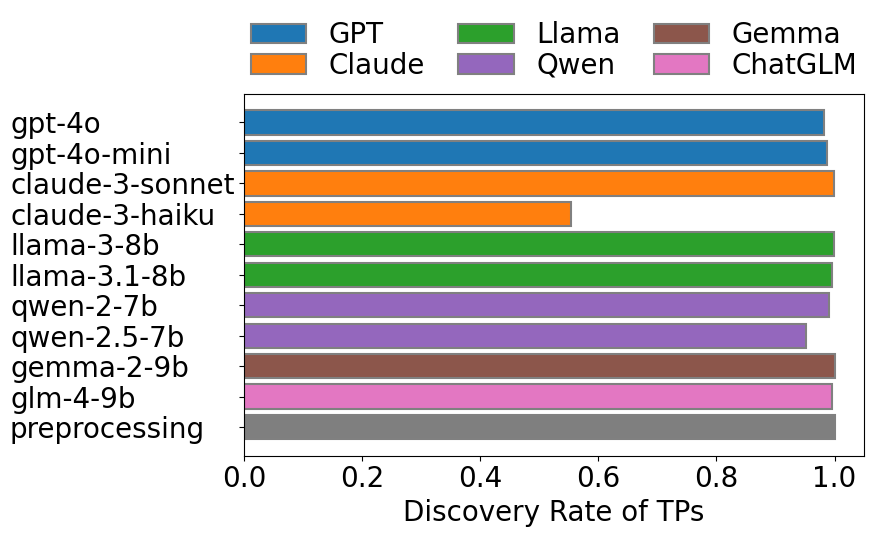}}        
\subfloat[PT4: WHOLE-FMA-SNOMED]{\includegraphics[width=0.33\textwidth]{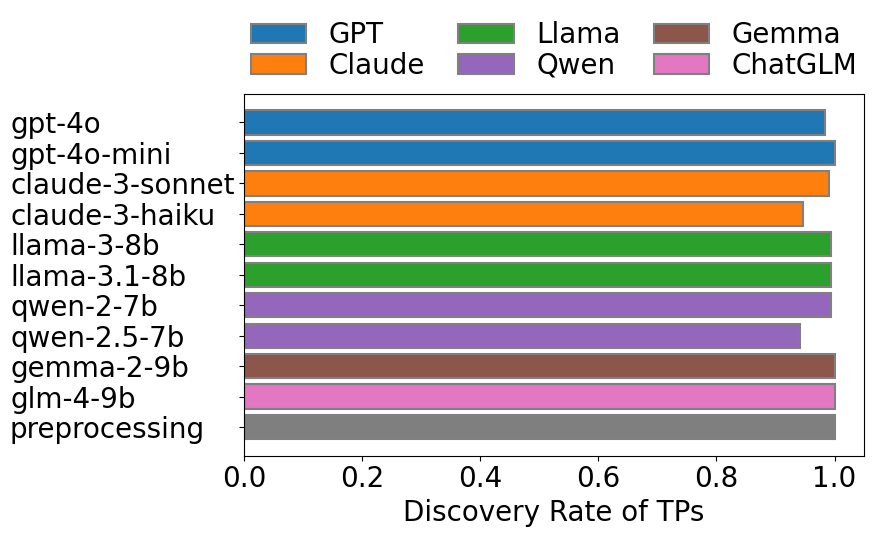}}            
\subfloat[PT4: WHOLE-SNOMED-NCI]{\includegraphics[width=0.33\textwidth]{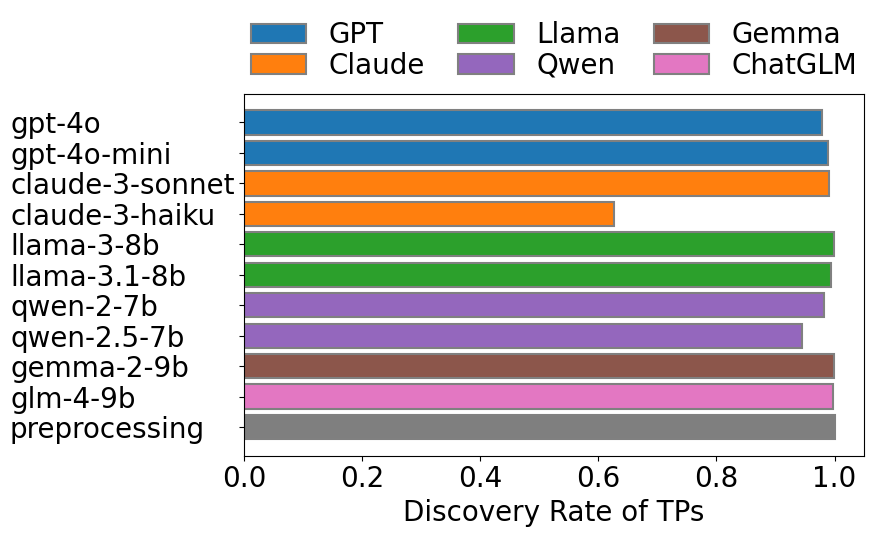}}   \\   
\caption{Discovery rate of TPs using LLMs on the {\itshape Largebio} Track.}
\label{fig: llm-tp}
\end{figure*}

Fig.~\ref{fig: llm-tp} compares LLMs with the classical text preprocessing pipeline in detecting TPs.  LLMs and the text preprocessing pipeline are expected to answer ``yes'' for TPs. The discovery rate is defined as the number of ``yes'' answers compared to the total number of TPs. For LLMs, we directly prompt LLMs to determine whether two entities are identical or not. For the text preprocessing pipeline, two entities are considered to be matched if they have the same results after the text preprocessing.  Here we use only Tokenisation (T) and Normalisation (N) to preprocess the text because Stop Words Removal (R) and Stemming/Lemmatisation (S/L) are shown to be less effective in our analysis in Section~\ref{sec: analysis}. Although some API-accessed commercial LLMs have shown the capability to detect TPs without a text preprocessing step, this behaviour varies across different LLMs and prompt templates. For different LLMs, open-source LLMs perform better than commercial API-accessed LLMs. It could be problematic for some commercial API-accessed LLMs (e.g. claude-3-haiku) to find two identical entities if text preprocessing is not applied prior to the comparison. They may consider entities with different formatting that are used in different contexts and therefore have different meanings. For different prompt templates, few-shot (PT2) significantly improves correctness, but self-reflection (PT3) may even hamper performance. The combination of few-shot and self-reflection (PT4) may even perform worse than the simple basic prompt (PT1). We believe that this phenomenon is mainly attributed to ``overthinking'', a type of LLM hallucination in which LLMs could over-analyse user input and generate unnecessarily complex details~\cite{chen2024not}. We also need to note here that the example generated in the few-shot prompt is based on our observation of the existing TPs. Finding a good example in real-world applications is not easy because ontology entities can have different formatting by nature. On the other hand, using the classical text preprocessing pipeline via function calling could help LLMs produce a stable result with a high discovery rate (almost 100\% from our observation) across different models and prompt templates.

\begin{figure*}[htbp]
\centering
\subfloat[PT1: WHOLE-FMA-NCI]{\includegraphics[width=0.33\textwidth]{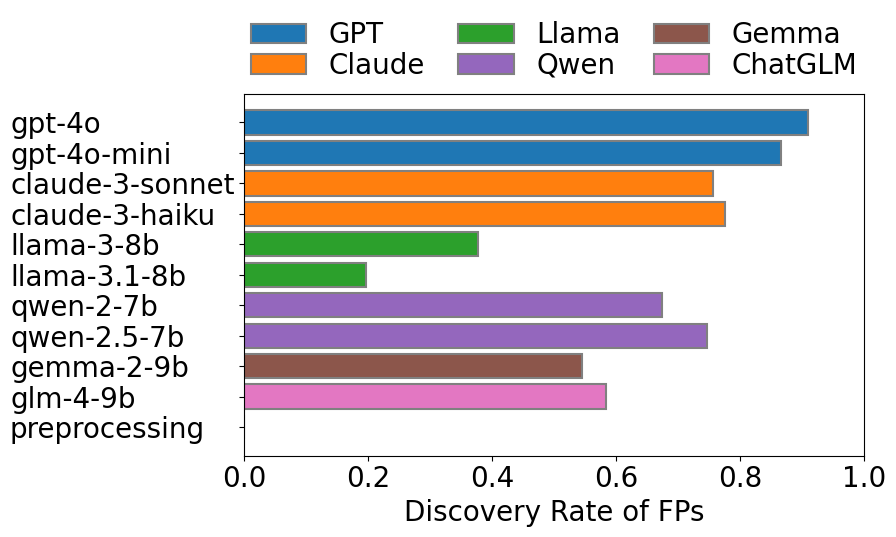}}
\subfloat[PT1: WHOLE-FMA-SNOMED]{\includegraphics[width=0.33\textwidth]{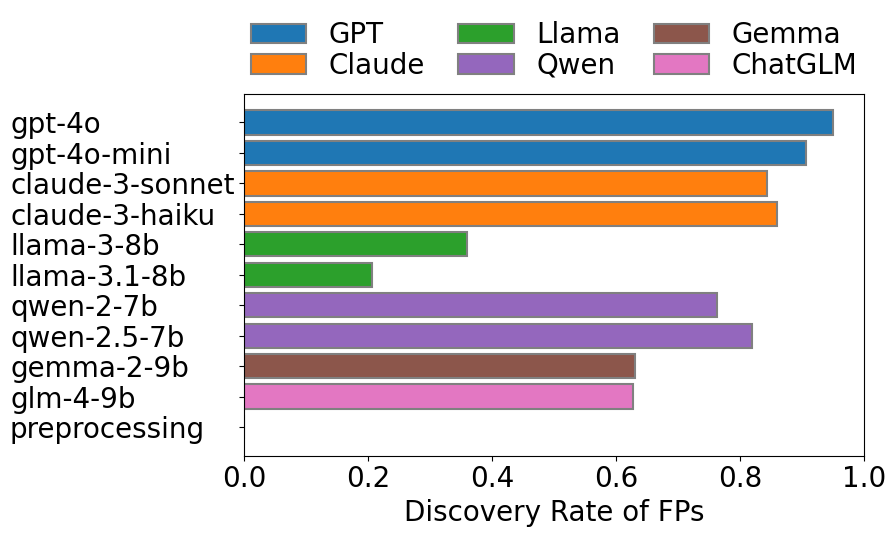}}
\subfloat[PT1: WHOLE-SNOMED-NCI]{\includegraphics[width=0.33\textwidth]{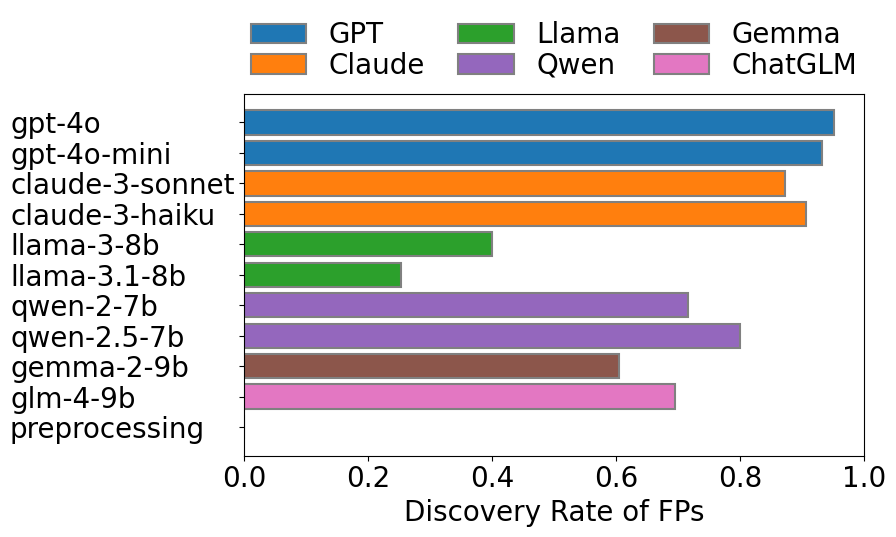}}                          \\
\subfloat[PT2: WHOLE-FMA-NCI]{\includegraphics[width=0.33\textwidth]{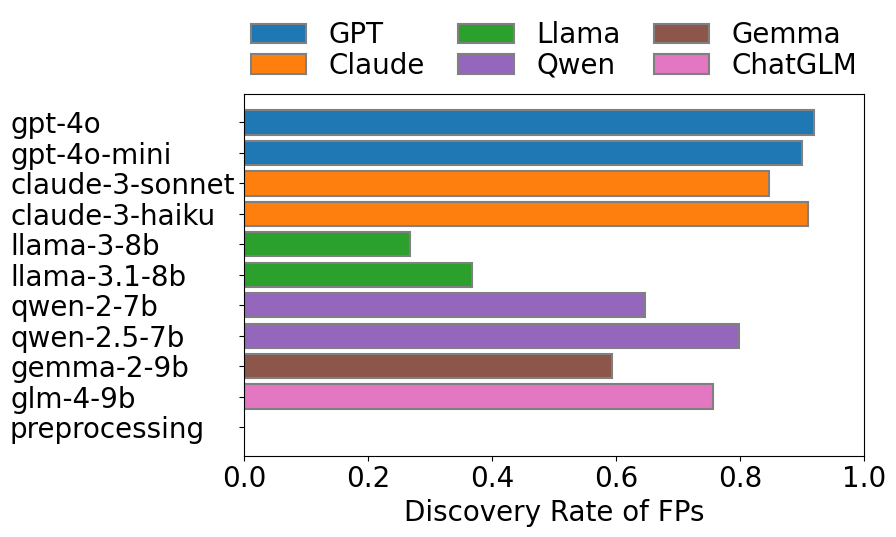}}
\subfloat[PT2: WHOLE-FMA-SNOMED]{\includegraphics[width=0.33\textwidth]{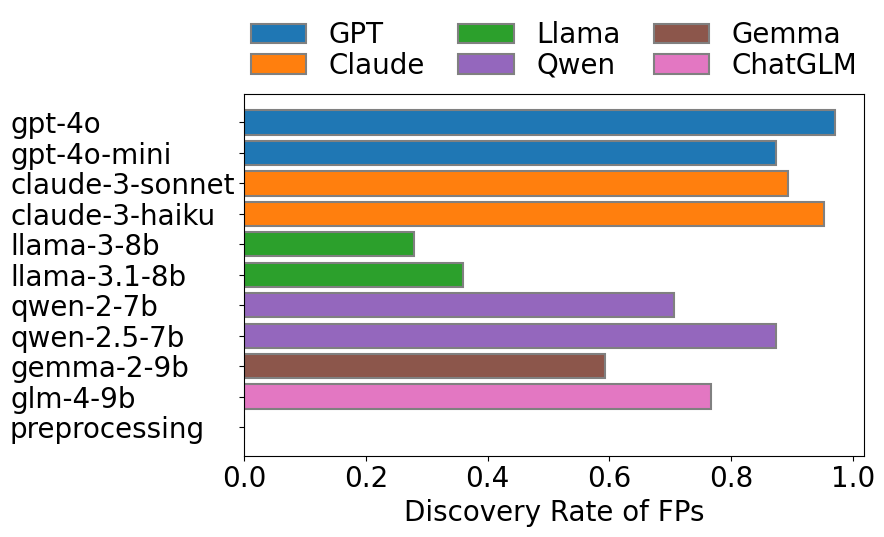}}
\subfloat[PT2: WHOLE-SNOMED-NCI]{\includegraphics[width=0.33\textwidth]{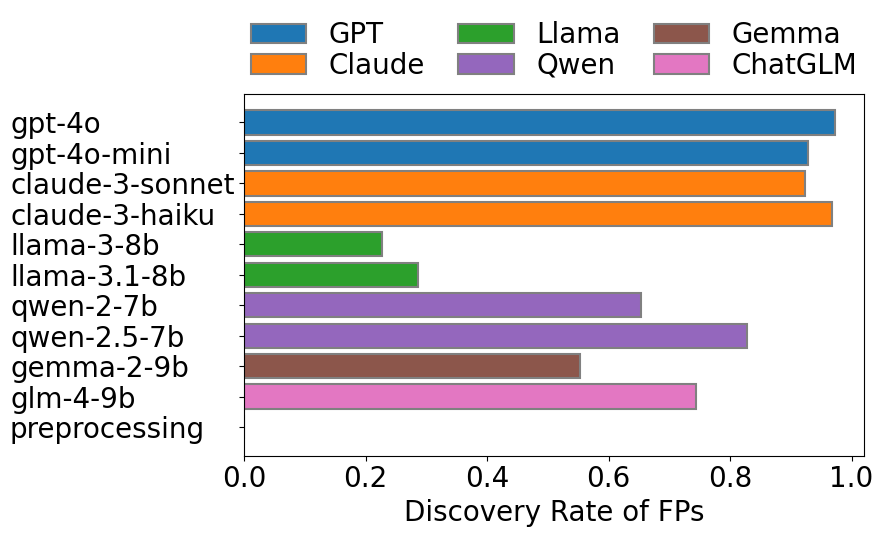}}                       \\
\subfloat[PT3: WHOLE-FMA-NCI]{\includegraphics[width=0.33\textwidth]{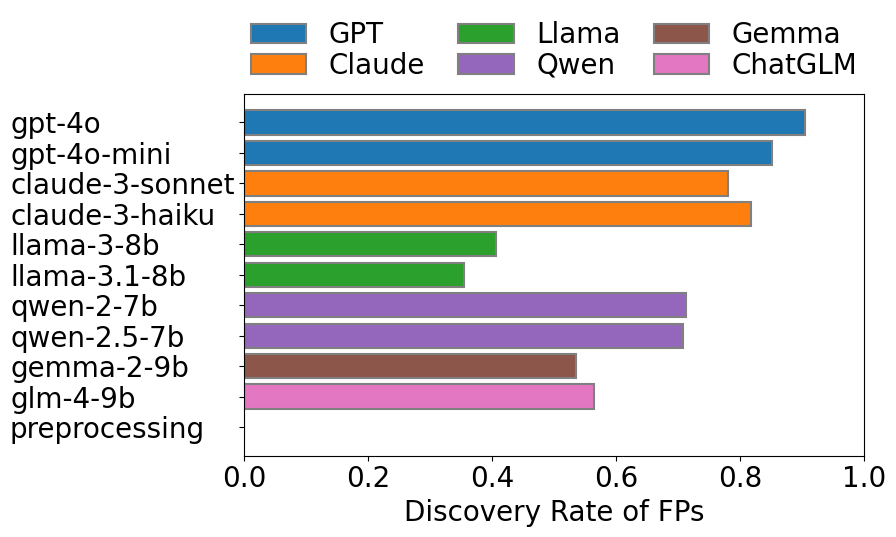}}
\subfloat[PT3: WHOLE-FMA-SNOMED]{\includegraphics[width=0.33\textwidth]{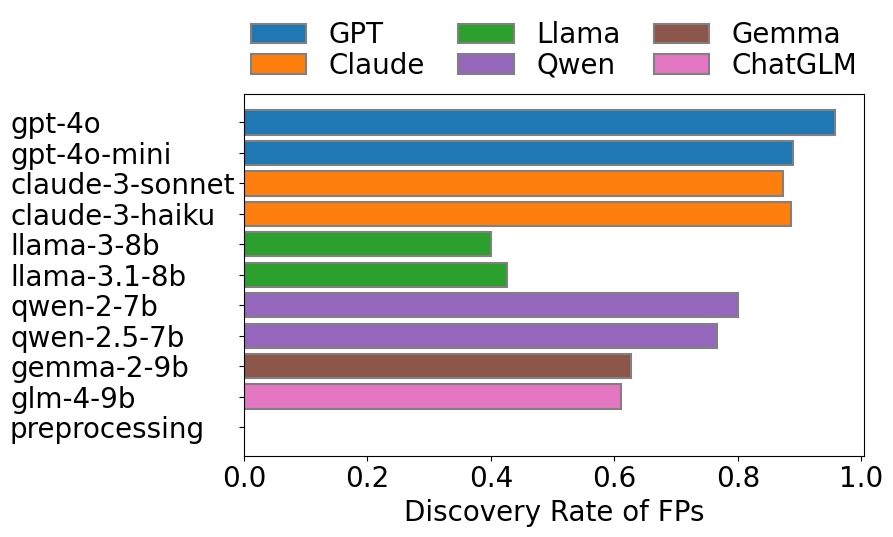}}
\subfloat[PT3: WHOLE-SNOMED-NCI]{\includegraphics[width=0.33\textwidth]{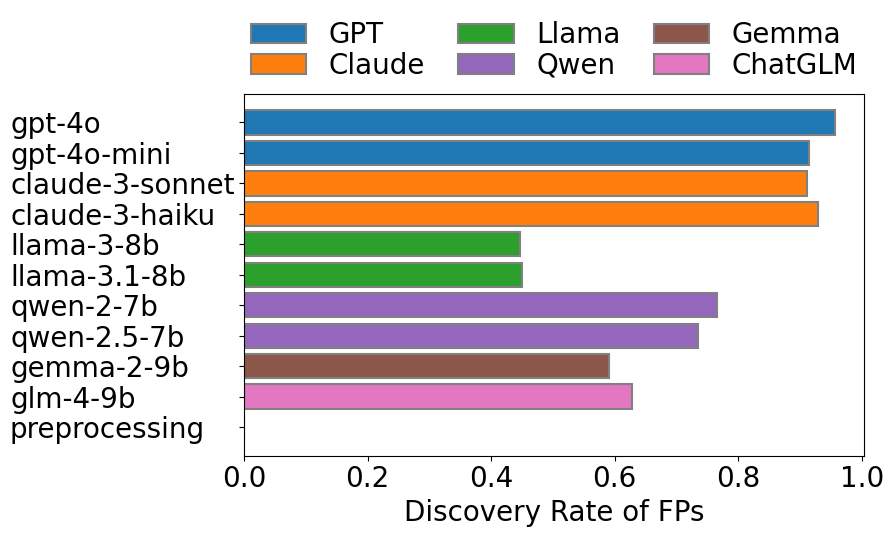}}                \\
\subfloat[PT4: WHOLE-FMA-NCI]{\includegraphics[width=0.33\textwidth]{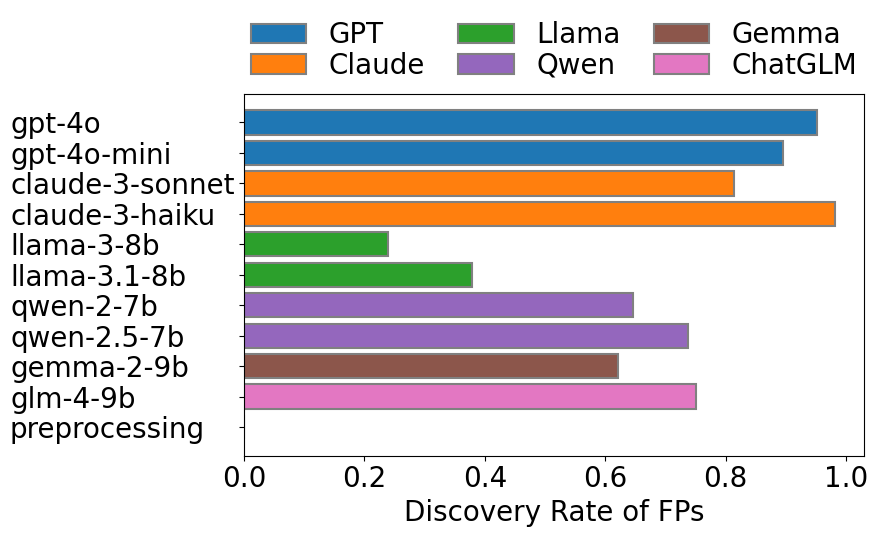}}       
\subfloat[PT4: WHOLE-FMA-SNOMED]{\includegraphics[width=0.335\textwidth]{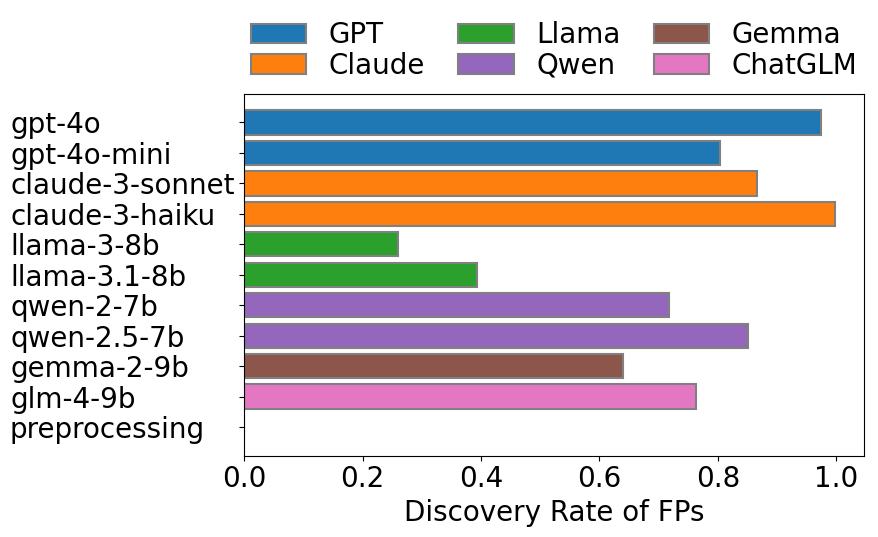}} 
\subfloat[PT4: WHOLE-SNOMED-NCI]{\includegraphics[width=0.33\textwidth]{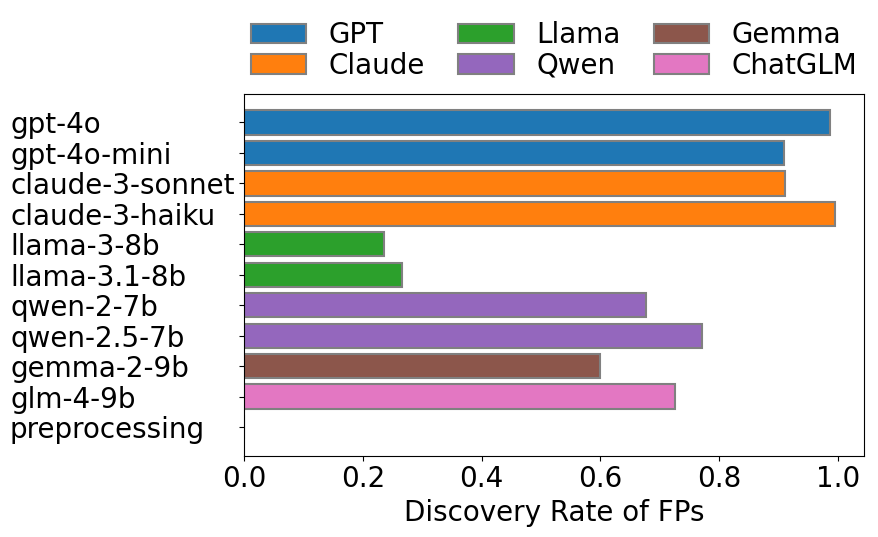}}   \\
\caption{The discovery rate of FPs using LLMs on the {\itshape Largebio} Track.}
\label{fig: llm-fp}
\end{figure*}

Fig.~\ref{fig: llm-fp} compares LLMs with the classical text preprocessing pipeline in detecting FPs. LLMs and the text preprocessing pipeline are expected to answer ``no'' for FPs. The discovery rate is defined as the number of ``no'' answers compared to the total number of FPs. Specifically, these FPs evaluated are non-trivial mappings that cannot be detected by the classical text preprocessing method. Here we use LLMs to detect FPs generated by the Lancaster Stemmer (SL). SL has shown a lower precision compared to other text preprocessing methods, which is caused by the method producing a significant number of FPs. LLMs have shown a strong capability to discover the FPs produced by the classical text preprocessing pipeline. For different models, API-accessed commercial LLMs are better than open-source LLMs and reach a higher discovery rate of more than 80\%. For different prompt templates, both few-shot (PT2), self-reflection (PT3), and a combination of the above (PT4) are better than a simple basic prompt (PT4). While mapping repair is usually non-trivial and requires domain expertise, LLMs have shown potential for performing a high-quality post hoc correction without human intervention.

For OM, we can see that LLMs are not good at detecting true mappings, but they are good at detecting false mappings. Therefore, we propose our post hoc LLM-based repair with two steps. (1) Given an alignment, employ the classical text preprocessing pipeline on each mapping via an LLM function calling (a.k.a. tool calling) to detect TPs. This means that the LLMs act as an interface, and the mapping judgement is based on the program logic defined in the text preprocessing method. This provides stable results for text preprocessing when finding FPs. LLM prompting is not suitable for this task, even with complex prompting strategies. (2) Given an alignment, use LLM prompting on each mapping to detect FPs. The prompt template can be simple like ``Is A equivalent to B?'', as we cannot see significant improvement using complex prompting strategies (except for Meta Llama models).

We apply our pipeline repair approaches on the Phase 2 text preprocessing method Lancaster Stemmer (SL). Due to the superior performance and superior  cost of the best LLMs, we evaluate only on the three largest datasets in the \textit{Largebio} Track: WHOLE-FMA-NCI, WHOLE-FMA-SNOMED, and WHOLE-SNOMED-NCI. With this choice, we expect the performance increase may justify the cost of the repair step.

\begin{figure}[htbp]
\centering
\subfloat{
\includegraphics[width=1\columnwidth]{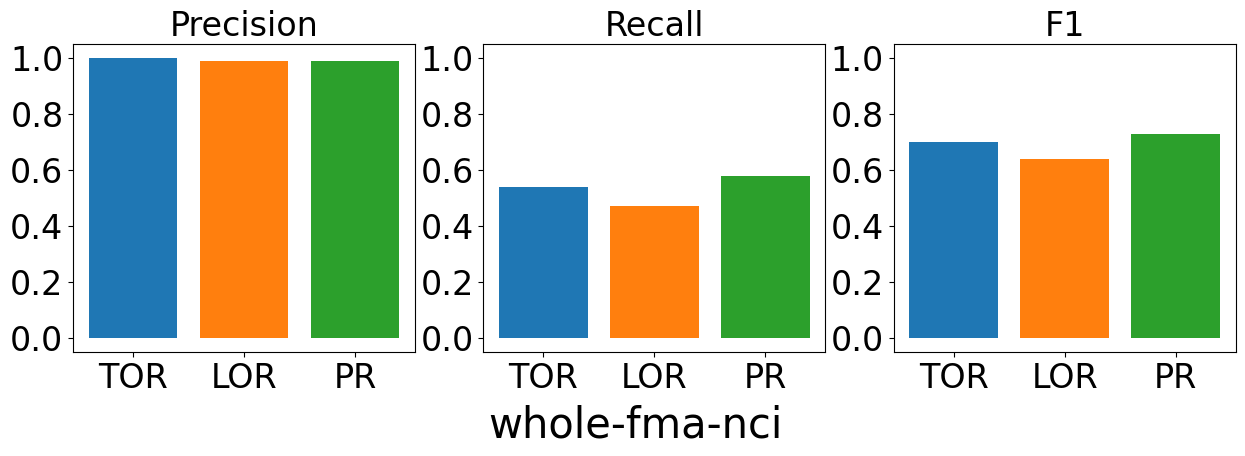}
}
\hfill
\subfloat{
\includegraphics[width=1\columnwidth]{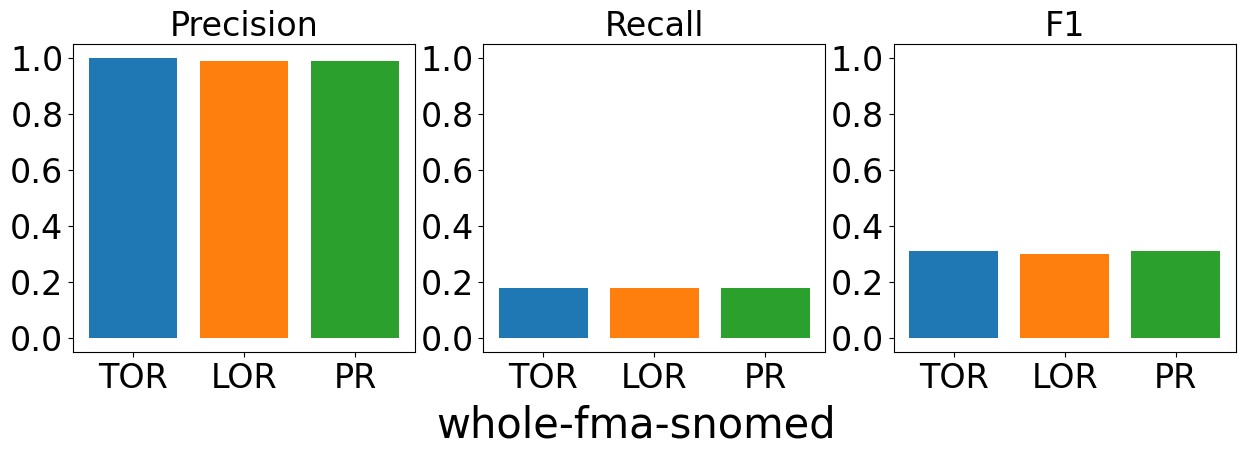}
}
\hfill
\subfloat{
\includegraphics[width=1\columnwidth]{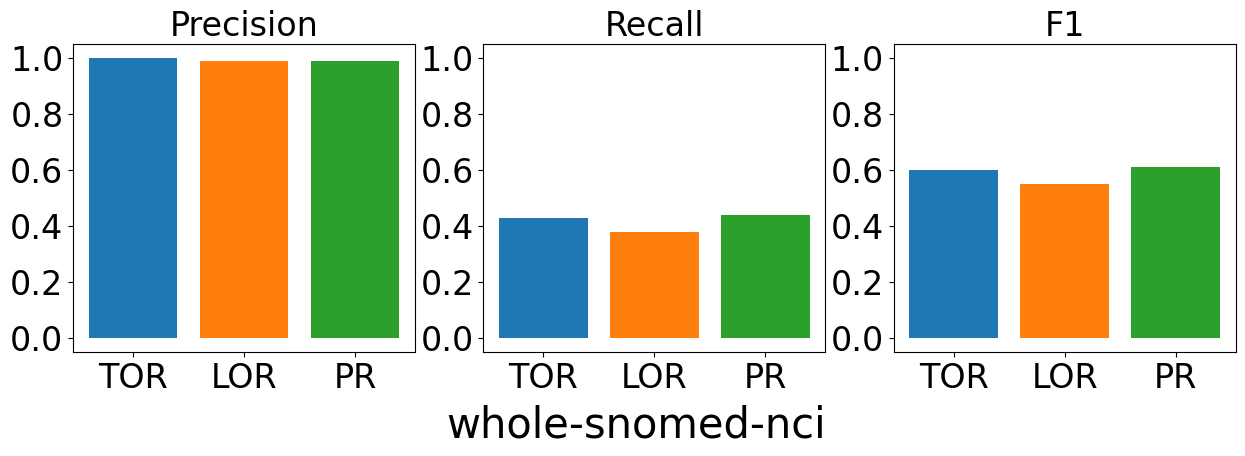}
}
\caption{Comparison of post hoc LLM-based repair (PR) with two alternatives: text-preprocessing-only repair (TOR) and LLM-only repair (LOR).}
\label{fig: pr-evaluation}
\end{figure}

Fig.~\ref{fig: pr-evaluation} compares post hoc LLM-based repair (PR) with two alternatives: text-preprocessing-only repair (TOR) and LLM-only repair (LOR). The results show that PR achieves almost the same precision as TOR and LOR, but its recall and F1 score are better than the other two alternatives. For OM, this means that our approach could detect more "seed mappings".

\begin{figure}[htbp]
\centering
\subfloat{
\includegraphics[width=1\columnwidth]{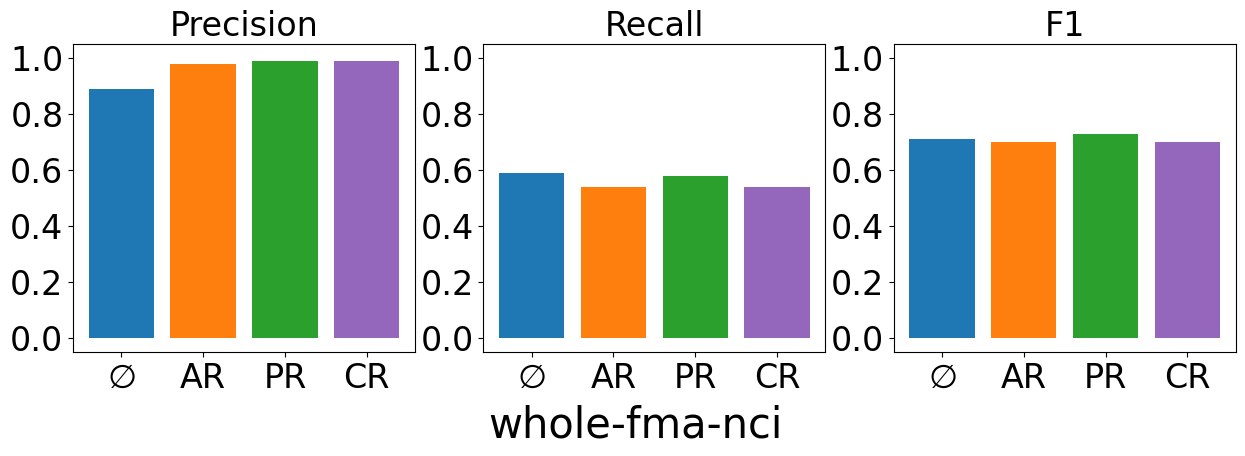}
}
\hfill
\subfloat{
\includegraphics[width=1\columnwidth]{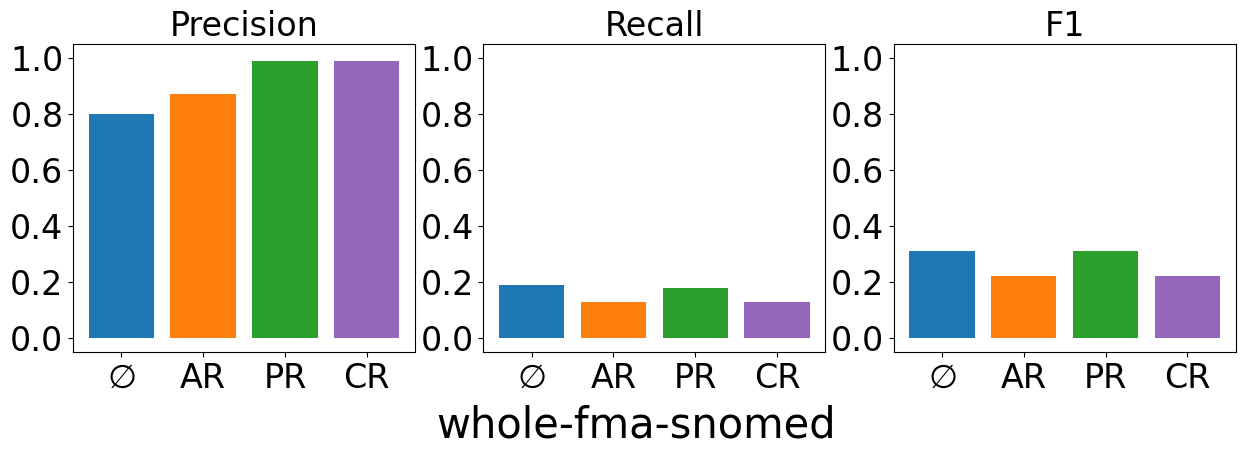}
}
\hfill
\subfloat{
\includegraphics[width=1\columnwidth]{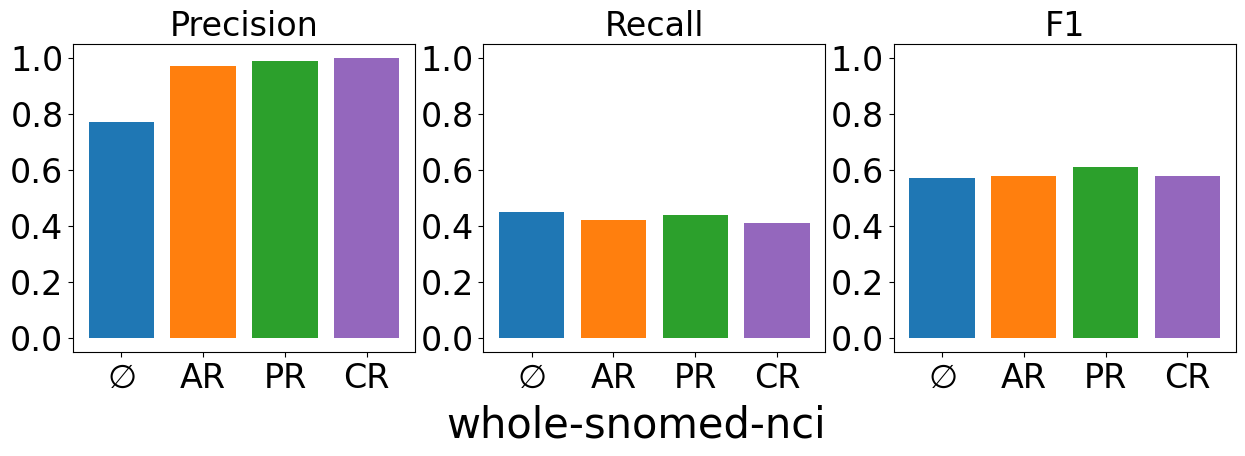}
}
\caption{Evaluation of pipeline repair approaches: Ad hoc logic-based repair (AR), post hoc LLM-based repair (PR), and the combination of above (CR). The symbol $\varnothing$ represents the baseline without text preprocessing repair.}
\label{fig: repair-evaluation}
\end{figure}

Fig.~\ref{fig: repair-evaluation} compares the matching performance with/without using the pipeline repair approaches: ad hoc logic-based repair (AR), post hoc LLM-based repair (PR), and the combination of the above (CR). We can see that AR and PR can significantly improve precision, with only a slight decrease in recall, but the overall F1 score also increases. For text preprocessing used in OM, its main purpose is to provide ``seed mappings'' for lexical and semantic matching. These mappings are meant to be precise and accurate. Both of our pipeline repair approaches shown with high precision are the best tool suites for OM. Although there is no restriction on the use of both ad hoc and post hoc repairs, the combination CR does not improve precision, recall, or F1 score.

\subsection{Evaluation within LLM-based Matchers}

We have shown that pre hoc logic-based repair improves syntactic matching. We have also shown that the post hoc repair method improves the detection of TPs and FPs. We now proceed to show how the repair methods perform in end-to-end modern LLM-based matchers. For this, we conduct an ablation study on one of the state-of-the-art OM systems: our Agent-OM system~\cite{qiang2023agent}. Agent-OM introduces a novel LLM-agent-based architecture for OM tasks. The system applies three levels of matching: syntactic, lexical, and semantic matching. The architecture contains two agents: a retrieval agent and a matching agent. Pre hoc logic-based repair and post hoc LLM-based repair have been applied to these agents respectively to improve system performance. For this ablation study, we compare Agent-OM that includes those repair components with a cut-down Agent-OM (Agent-OM-Minus) without these repair components.

Pre hoc logic-based repair is used in the syntactic retriever of the retrieval agent. While Agent-OM uses T+N to process the entity name, it has limited capability to capture micro-level similarity, such as ``Air\_Handler\_Unit'' and ``Air\_Handling\_Unit''. Using R and S/L can handle these cases but leads to more FPs, which pre hoc logic-based repair can resolve. Post hoc LLM-based repair is used as a validator to filter the matching candidates produced by the matching agent. While this approach may also filter a small number of TPs, we expect the benefits to outweigh the drawbacks.

\section{Limitations}
\label{sec: limitations}

The references contained in the OAEI tracks are not perfect gold standards and some need further development. Therefore at present 100\% accuracy is not necessarily desirable. In this study, we deal only with equivalence mappings between classes and properties. Subsumption mappings are not considered. Finding widely used baseline data with ground truth matches for subsumptions is empirically difficult. Only a few OAEI tracks contain subsumption mappings, and they are not enough to robustly demonstrate the text preprocessing used in subsumption mappings.

\section{Conclusion}
\label{sec: conclusion}

In this paper, we conduct a comprehensive study on the effect of the text preprocessing pipeline on OM. 8 OAEI tracks with 49 distinct alignments are evaluated. Despite the importance of text preprocessing in OM, our experimental results indicate that the text preprocessing pipeline is currently ill-equipped. We find that Phase 1 text preprocessing methods (Tokenisation and Normalisation) help with both matching completeness (i.e. recall) and correctness (i.e. precision). Phase 2 text preprocessing methods (Stop Words Removal and Stemming/Lemmatisation) are less effective. They can improve matching completeness (i.e. recall), but matching correctness (i.e. precision) is relatively low. We propose two novel pipeline repair methods to repair the less effective Phase 2 text preprocessings. Both approaches outperform the traditional text preprocessing pipeline without repair, in particular, the matching correctness (i.e. precision) and overall matching performance (i.e. F1 score). We argue that the traditional text preprocessing pipeline is not obsolete in the era of LLMs. It is an integral part of the LLMs used in OM tasks. Integrating the classical text preprocessing pipeline can significantly improve the robustness of purely prompt-based approaches.

While the text preprocessing pipeline in OM is usually applied on the basis of intuition or extrapolation from other experiences, this work advances the state of knowledge towards making design decisions objective and supported by evidence.

\begin{itemize}[wide, noitemsep, topsep=0pt, labelindent=0pt]
\item Our study shows (a) \emph{whether} to use, or not use, and (b) \emph{how} to use these text preprocessing methods. Such experimental results will benefit the selection of appropriate text preprocessing methods for OM. For large-scale OM, it can significantly reduce unnecessary trial costs.
\item Our pipeline repair approaches can be used to repair the less effective Phase 2 text preprocessing methods. Its broader value is showing how to maximise true mappings and minimise false mappings throughout the text preprocessing pipeline. From a statistical perspective, it is a local optimisation for syntactic matching that will benefit the global optimisation for OM.
\item We showcase the opportunities for integrating the text preprocessing pipeline with LLMs in OM. We find that LLMs are good tools to find false mappings, but their outputs for finding true mappings are non-deterministic. The best practice is to integrate the classical text preprocessing pipelines with the LLMs. Using function calling to invoke the classical text preprocessing pipeline within LLMs, our approach could produce more stable and robust ``seed mappings’’ than traditional text preprocessing methods and purely prompt-based approaches.
\item Our study demonstrates that the nature of text preprocessing used in downstream applications can be context-based. Not all methods are effective, and some of them may even hamper the performance of downstream tasks. Modifications are required before applying text preprocessing methods to specific tasks.
\end{itemize}

Our future work will focus on handling class axioms and complex relationships to evaluate the text preprocessing pipeline. We will also study the pipeline and pipeline repair approach working with both knowledge-based OM systems and LLM-based OM systems. This work has been included in the Agent-OM entry to the 2025 OAEI campaign.

\section*{Artifacts}
\label{sec: artifacts}

The source code, data, and/or other artifacts have been made available at \url{https://github.com/qzc438-research/ontology-nlp}. The OAEI datasets are available from the OAEI Matching EvaLuation Toolkit (MELT) at \url{https://dwslab.github.io/melt/track-repository} (retrieved January 1, 2025). The OAEI data policy can be found in \url{https://oaei.ontologymatching.org/doc/oaei-deontology.2.html}.

\section*{Acknowledgments}

The authors thank Sven Hertling for curating the OAEI datasets. The authors thank Ernesto Jimenez-Ruiz for helpful advice on reproducing the OAEI Largebio Track. The authors thank Alice Richardson of the Statistical Support Network (Australian National University) for helpful advice on the statistical analysis in this paper. The authors thank the Commonwealth Scientific and Industrial Research Organisation (CSIRO) for supporting this project.

\bibliographystyle{IEEEtran}
\balance
\bibliography{mybibliography}

\end{document}